\documentclass[conference]{IEEEtran}
\IEEEoverridecommandlockouts
\usepackage{amsmath,amssymb,amsfonts}
\usepackage{algorithmic}
\usepackage{graphicx}
\usepackage{textcomp}
\usepackage{xcolor}
\usepackage{booktabs}
\usepackage{multirow}
\usepackage{algorithm}
\usepackage[colorlinks=false, linkcolor=blue, urlcolor=blue, citecolor=green]{hyperref}
\usepackage{cite}

\begin{document}

\title{Threshold Modulation for Online Test-Time Adaptation of Spiking Neural Networks\\

\thanks{\textsuperscript{*} Corresponding authors}
}

\author{\IEEEauthorblockN{Kejie Zhao\textsuperscript{1}, Wenjia Hua\textsuperscript{1}, Aiersi Tuerhong\textsuperscript{2}, Luziwei Leng\textsuperscript{3}, Yuxin Ma\textsuperscript{1,*}, and Qinghai Guo\textsuperscript{3,*}}
\IEEEauthorblockA{
\textsuperscript{1}\textit{Department of CSE, Southern University of Science and Technology}, Shenzhen, China\\
\textsuperscript{2}\textit{College of Mathematics and Statistics, Chongqing University}, Chongqing, China\\
\textsuperscript{3}\textit{ACS Laboratory, Huawei Technologies Co., Ltd.}, Shenzhen, China\\
\{zhaokj2023, huawj2023\}@mail.sustech.edu.cn, 20211385@stu.cqu.edu.cn, lengluziwei@huawei.com,\\
mayx@sustech.edu.cn, guoqinghai@huawei.com}
}

\maketitle

\begin{abstract}
Recently, spiking neural networks (SNNs), deployed on neuromorphic chips, provide highly efficient solutions on edge devices in different scenarios. However, their ability to adapt to distribution shifts after deployment has become a crucial challenge. Online test-time adaptation (OTTA) offers a promising solution by enabling models to dynamically adjust to new data distributions without requiring source data or labeled target samples. Nevertheless, existing OTTA methods are largely designed for traditional artificial neural networks and are not well-suited for SNNs. To address this gap, we propose a low-power, neuromorphic chip-friendly online test-time adaptation framework, aiming to enhance model generalization under distribution shifts. The proposed approach is called Threshold Modulation (TM), which dynamically adjusts the firing threshold through neuronal dynamics-inspired normalization, being more compatible with neuromorphic hardware. Experimental results on benchmark datasets demonstrate the effectiveness of this method in improving the robustness of SNNs against distribution shifts while maintaining low computational cost. The proposed method offers a practical solution for online test-time adaptation of SNNs, providing inspiration for the design of future neuromorphic chips. The demo code is available at \href{https://github.com/NneurotransmitterR/TM-OTTA-SNN}{github.com/NneurotransmitterR/TM-OTTA-SNN}.

\end{abstract}



\section{Introduction}
In recent years, with the rapid development of high-performance hardware and training algorithms, modern deep artifical neural networks (ANNs) can have billions, or even hundreds of billions, of parameters, requiring large-scale computational resource for training and inference. Despite the impressive performance of ANNs, the application bottlenecks caused by their high energy consumption are becoming increasingly evident, especially in scenarios that demand low latency and low power. Spiking Neural Networks (SNNs), a brain-inspired neural network model, have regained attention from researchers in recent years. Unlike traditional neural networks, SNNs transmit information through discrete spike events, and their inherent sparsity enables efficient information transmission with much lower power consumption. This makes them particularly suitable for neuromorphic computing. In recent years, researchers have made significant progress in learning rules and network architectures \cite{huLargescaleSpikingNeural2024}, enabling deep SNNs to achieve performance comparable to ANNs on certain tasks while consuming much less energy when deployed on neuromorphic chips in edge devices.

For pre-trained models deployed on edge devices, test-time adaptation (TTA) plays a crucial role in addressing data distribution shifts by dynamically adjusting the model, enabling improved performance on out-of-distribution test data without additional labels or offline training. However, deploying TTA algorithms on edge devices must overcome constraints such as limited computational capacity, energy sensitivity, and the need for real-time operation with hardware compatibility. With their event-driven and sparse computation characteristics, SNNs have emerged as an ideal choice for low-power edge computing, but like any neural network, their performance can also significantly degrade in scenarios involving input distribution shifts like environmental changes or sensor aging. Therefore, designing energy-efficient TTA algorithms to enhance the adaptability of on-chip SNNs to dynamic environments not only improves their robustness but also meets the resource constraints of edge devices, advancing the broader adoption of low-power intelligent systems. Regarding recent works, SNNTL \cite{zhanEffectiveTransferLearning2022} proposed a transfer learning framework for SNNs; however, it requires labeled samples from both the target and source domains, which is not accessible in fully test-time adaptation settings. Duan et al. \cite{duanBrainInspiredOnlineAdaptation2025} were the first to address online adaptation for remote sensing with SNNs converted from ANNs, but the method's feasibility for direct adaptation on neuromorphic chips is limited. In this paper, we propose an online test-time adaptation framework for SNNs, specifically designed for on-chip online adaptation scenarios. This framework does not require access to source data or target data labels and is more compatible with online on-chip learning in neuromorphic chips.

This work makes the following key contributions: (1) We propose a low-power online test-time adaptation framework for SNNs, which is one of the first works specifically addressing this issue.  
(2) We examine normalization calibration and entropy minimization in online test-time adaptation for SNNs, and explore optimal configurations via an ablation study. (3) The proposed Threshold Modulation module enables models to adapt without introducing significant overhead in a chip-friendly way, inspiring the design of future neuromorphic chips.

\section{Related Work}

\subsection{Spiking Neural Networks}
Spiking Neural Networks (SNNs) are a type of neural networks that mimics biological neuron activity by transmitting information through binary spikes, rather than continuous activations. Each neuron has a membrane potential that is influenced by incoming spikes, and when it reaches a threshold, the neuron emits a spike and resets. This event-driven nature of SNNs make them well-suited for low-power applications. Equations~\eqref{eq:lif-charge}, \eqref{eq:lif-fire} and \eqref{eq:lif-reset} present an iterative hard-reset version of the commonly used Leaky Integrate-and-Fire (LIF) neuron model \cite{fangDeepResidualLearning2021} 
which maintains biological plausibility while achieving high computational efficiency. In the equations, $t$ denotes the simulation time step ($\textstyle{1\le t \le T}$), $X_{t}$ represents the input current, $h_{t}$ is the membrane potential after charging, $u_{t}$ is the membrane potential after firing, $\tau$ is the decay constant, $V_{th}$ is the firing threshold, $\Theta$ is the Heaviside step function, and $o_{t}$ is the spike output. The neuronal dynamics can be divided into three parts: neuronal charging \eqref{eq:lif-charge}, neuronal firing \eqref{eq:lif-fire}, and neuronal reset \eqref{eq:lif-reset}.
\begin{align}
    h_{t} &=X_{t}+1/\tau\cdot u_{t-1} \label{eq:lif-charge} \\
    o_{t} &=\Theta(h_{t}-V_{th}) \label{eq:lif-fire} \\
    u_{t} &=h_{t}\cdot (1-o_{t})+o_{t}\cdot V_{reset} \label{eq:lif-reset}
\end{align}

In deep learning research, two commonly adopted approaches to train SNNs are ANN-SNN conversion \cite{rueckauerConversionContinuousValuedDeep2017a} and direct training using Backpropagation Through Time with surrogate gradients (BPTT with SG) \cite{wuSpatioTemporalBackpropagationTraining2018,neftciSurrogateGradientLearning2019}. The first approach utilizes pre-trained ANNs and convert the model weights, while the second approach overcomes the discontinuous nature of the spike function by approximating its gradients, allowing less training time and greater flexibility.

Meanwhile, SNNs are closely linked to neuromorphic chips, which provide hardware support for their practical use. Neuromorphic chips simulate neuron behavior and synaptic connections, significantly improving the energy efficiency and computational power of SNNs. Chips like Loihi \cite{daviesLoihiNeuromorphicManycore2018a}, TrueNorth \cite{akopyanTruenorthDesignTool2015a}, Speck \cite{yaoSpikebasedDynamicComputing2024} and Darwin3 \cite{maDarwin3LargescaleNeuromorphic2024} provide the hardware foundation for the practical use of SNNs, advancing brain-inspired computing. Therefore, when developing SNN algorithms, it is best to consider their potential for implementation on neuromorphic hardware in order to fully harness the capabilities of SNNs.

\subsection{On-chip learning on neuromorphic chips}

The aforementioned neuromorphic chips are initially designed solely for inference, to reveal its extreme efficiency at the edge. Recently, on-chip learning on those chips has gained more research interest as it provides an on-the-fly adaptation to changes in the environment. 

A biologically-plausible approach for on-chip learning on neuromorphic chips is to utilize the local synaptic plasticity. For instance, ROLLS\cite{qiao2015reconfigurable} proposed to use spike-driven synaptic plasticity (SDSP) to perform on-chip learning for simple classification tasks on a tiny chip with $256$ analog neurons and $128k$ synapses. ODIN\cite{frenkel20180} extended the SDSP method into digital chips with a similar scale. Loihi\cite{daviesLoihiNeuromorphicManycore2018a} and its following updated version generalized this type of learning rule into a more general programmable synaptic plasticity rule to support different scenarios. These methods, though quite biologically-plausible and highly efficient, lack the ability to learn short-to-long-term temporal dependencies and thus their usability in more complex tasks is restricted.

Another approach is to develop a hardware friendly implementation of Backpropagation Through Time (BPTT), a quite popular deep learning framework for spatial-temporal neural networks like SNNs. Among them, e-prop\cite{bellecSolutionLearningDilemma2020} is a widely used framework, which has been revised to support both small scale chips such as ReckOn\cite{frenkel2022reckon} and large scale chips such as SpiNNaker\cite{rostami2022prop}, showing their effectiveness on different machine learning tasks. However, such an approximation of BPTT still relies on labeled samples or requires multiple epochs of learning stage, hence is not an efficient way for online test-time adaptation tasks. In fact, current on-chip learning implementations are primarily designed for end-to-end learning and have not been developed for test-time adaptation scenarios, limiting their usability.

\begin{figure*}[!t]
    \centering
    \begin{minipage}{\linewidth}
        \centering
        \includegraphics[width=\linewidth]{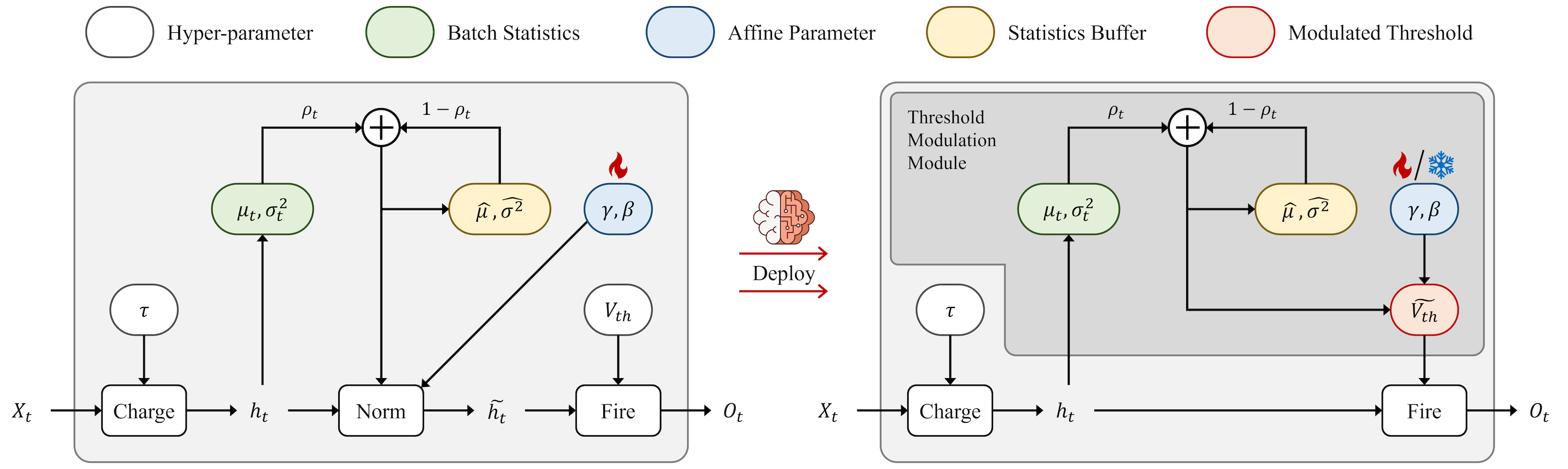}
    \end{minipage}%
    \vspace{2pt}
    \begin{minipage}{0.5\linewidth}
        \centering
        {\fontsize{8pt}{10pt}\selectfont
        (a) Pre-train}
    \end{minipage}%
    \begin{minipage}{0.5\linewidth}
        \centering
        {\fontsize{8pt}{10pt}\selectfont
        (b) Test time}
    \end{minipage}%
    \vspace{-6pt}
    \caption{\textbf{Framework overview.} \textbf{(a)} In the pre-training phase, membrane potentials are normalized after neuronal charging and affine parameters are trained. Before test-time adaptation, the model can be mapped and deployed on a neuromorphic chip after threshold re-parameterization. \textbf{(b)} During adaptation on-chip, statistics are updated to modulate the firing threshold. The affine parameters can also be optimized upon need, resulting in two main variants of our method: \textbf{TM-ENT} and \textbf{TM-NORM}.}
    \label{fig:method}
    \vspace{-10pt}
\end{figure*}

\subsection{Online Test-Time Adaptation}

Test-Time Adaptation (TTA) is a framework designed to address distribution shifts between training and testing data, enabling pre-trained models to adapt dynamically to unlabeled target data during inference. Unlike traditional domain adaptation methods, TTA operates without access to source data during test time, making it highly applicable in resource-constrained scenarios.

Online Test-Time Adaptation (OTTA) extends TTA to streaming data scenarios, allowing models to dynamically adapt to sequentially arriving data. OTTA incorporates knowledge from previously seen data to iteratively refine the model. Normalization calibration \cite{wangTentFullyTestTime2020,hongMECTAMemoryEconomicContinual2022}, entropy minimization \cite{wangTentFullyTestTime2020,niuStableTesttimeAdaptation2022}, pseudo-labeling \cite{boudiafParameterfreeOnlineTesttime2022,jangTestTimeAdaptationSelfTraining2022}, and teacher models \cite{wangContinualTestTimeDomain2022} are commonly employed. On the other hand, anti-forgetting mechanisms \cite{wangContinualTestTimeDomain2022,niuEfficientTestTimeModel2022} address the degradation of source domain performance during online adaptation. OTTA’s ability to handle dynamic and evolving distributions makes it particularly suited for real-time applications in complex and shifting environments.

Although remarkable progress has been made in OTTA in recent years, there is a notable lack of methods specifically proposed for SNNs. Existing OTTA methods are largely designed for ANNs and face various challenges when directly applied to SNNs deployed on neuromorphic chips, such as immutable weights, inputs, and outputs. Reference \cite{duanBrainInspiredOnlineAdaptation2025} was the first to address the online adaptation for remote sensing with SNNs. Based on entropy minimization, they utilize an online learning algorithm and adaptive activation scaling to accelerate SNN adaptation. However, the update of statistics and affine parameters in normalization layers prevents these layers from being fused into the model weights or neuron parameters, increasing both the computational cost and the complexity of on-chip implementation.
\section{Method}

Given the existing work and the lack of OTTA algorithms specifically designed for SNNs, the online test-time adaptation framework we propose aims to achieve a neuromorphic hardware-friendly, low-power, and efficient framework for online test-time adaptation. In this chapter, we present the details of this framework.

\subsection{Overview of the Test-time Adaptation Framework}

Fig.~\ref{fig:method} provides an overview of the proposed TTA framework for SNNs based on the Threshold Modulation (TM) module. Our approach integrates three phases: pre-training, deployment, and online adaptation. The key idea is to minimize the additional computational cost while performing the adaptation within the neuron rather than at the input. Pre-training enables the model to learn features on the source domain; after deployment, it can perform online adaptation using the Threshold Modulation module, allowing the deployed model to achieve online test-time adaptation through neuron-level operations without modulating the normalization layers, mutating the model weights or modifying the output.

\subsection{Membrane Potential Batch Normalization}
Batch normalization (BN)\cite{ioffeBatchNormalizationAccelerating2015} is widely used in training deep neural networks, as it helps reduce internal covariate shift and facilitates model convergence. According to pioneering TTA research \cite{nadoEvaluatingPredictionTimeBatch2021,schneiderImprovingRobustnessCommon2020}, severity of covariate shift correlates with performance degradation, so calibating the BN statistics alleviates the degradation by removing covariate shift. Meanwhile, entropy minimization \cite{wangTentFullyTestTime2020} is used to boost the performance by updating the affine parameters of BN. BN calibration and entropy minimization constitute the fundamental methods of OTTA. However, directly calibrating the statistics or updating the affine parameters not only requires a significant amount of computation but, most importantly, is incompatible with current neuromorphic chip designs and network mapping methods. For example, in recent studies, the BN layers in pre-trained convolutional spiking neural networks are converted into model weights \cite{rueckauerConversionContinuousValuedDeep2017a} or leakage terms of neuron input \cite{esserConvolutionalNetworksFast2016} to facilitate on-chip implementation.

Such infeasibility prompts us to consider normalization operations on the neuronal membrane potential rather than the input. The Membrane Potential Batch Normalization (MPBN) method proposed in \cite{guoMembranePotentialBatch2023a} aligns well with our requirements and can be applied to the pre-training phase of the framework. MPBN modifies the neuronal dynamics by performing batch normalization on the membrane potential after neuronal charging. Equations~\eqref{eq:mpbn-norm} and \eqref{eq:mpbn-fire} illustrates the modification of neuronal dynamics by MPBN: the membrane potential is batch-normalized before neuronal firing. 
\begin{align}
    \tilde{h_{t}} &=\mathbf{BN}(h_{t}) \label{eq:mpbn-norm} \\
    o_{t}& =\mathbf{\Theta}(\tilde{h_{t}}-V_{th}) \label{eq:mpbn-fire}
\end{align}

Furthermore, by unfolding the MPBN, we can obtain the equivalent firing threshold while eliminating the MPBN during inference, referred to as threshold re-parameterization  \cite{guoMembranePotentialBatch2023a}.
\begin{align}
    &o_{t,i} =
        \begin{cases}
         1 & \text{ if } \gamma_{i}\frac{u_{t,i}-\mu_{i}}{\sqrt{\sigma_{i}^2}}+\beta_{i} > V_{th} \\
         0 & \text{ otherwise }
        \end{cases} \\
    &o_{t,i} =
        \begin{cases}
         1 & \text{ if } u_{t,i}>\frac{(V_{th}-\beta_{i})\sqrt{\sigma_{i}^{2}}}{\gamma_{i}}+\mu_{i} \\
         0 & \text{ otherwise }
        \end{cases} \\
    &(\tilde{V}_{th})_{i} =\frac{(V_{th}-\beta_{i})\sqrt{\sigma_{i}^{2}}}{\gamma_{i}}+\mu_{i}
    \label{eq:trp}
\end{align}

According to \eqref{eq:trp}, neuronal firing can now use the new threshold $\tilde{V}_{th}$, where $\mu$ and $\sigma^2$ are the mean and the variance, $\gamma$ and $\beta$ are the learnable parameters. During deployment, MPBN can be fused into new thresholds. What was not addressed in the original paper is that, technically, merely altering the threshold does not ensure equivalence in inference when the simulation time steps $T>1$. To achieve complete equivalence, it is necessary to perform additional normalization of the membrane potentials of non-firing neurons using the current statistics. In the proposed Threshold Modulation module, we reconsider the inclusion of this operation as an optional component and experimentally validate its impact by ablation study.

\subsection{Threshold Modulation Module}
\label{subsec:tm}
Indeed, MPBN along with threshold re-parameterization enable the adaptation to be applied at the neuronal threshold level rather than to BN statistics or parameters, making it more feasible for on-chip implementation. Based on this observation, we propose a method that integrates pre-training, deployment, and on-chip test-time adaptation, referred to as Threshold Modulation (TM). The Threshold Modulation module is an additional modification to the firing phase of the spiking neuron. For a pre-trained model with learnable thresholds \cite{wangLTMDLearningImprovement2022,rathiDIETSNNLowLatencySpiking2023}, it can also be modulated during test time to achieve test-time adaptation after the learned threshold being the new $V_{th}=V_{th}^{pre}$ during test-time.
\begin{equation}
    \hat{\mu}=(1-\rho_{t})\cdot\hat{\mu}+\rho_{t}\cdot\mu_{t}
    \label{eq:mu}
\end{equation}
\begin{equation}
    \hat{\sigma^2}=(1-\rho_{t})\cdot\hat{\sigma^2}+\rho_{t}\cdot\sigma^{2}_{t}
    \label{eq:sigma2}
\end{equation}
\begin{equation}
    \widetilde{V_{th}} = {(V_{th}-\beta)\cdot\sqrt{\hat{\sigma^{2}}}}/{\gamma} + \hat{\mu}
    \label{eq:Vth}
\end{equation}

Equations \eqref{eq:mu} and \eqref{eq:sigma2} demonstrate the exponential moving average of the estimated statistics of the charged membrane potentials. Equation \eqref{eq:Vth} defines the modulated firing threshold at time step $t$.

\begin{figure}[t]
    \vspace{-10pt}
    \begin{algorithm}[H]
    \caption{Adaptation of one LIF layer using TM}
        \textbf{Hyper-params:} $V_{th}$, $V_{reset}$, $\tau$, $\omega$, $\rho_{0}$, $r\in\{0,1\}$, $e\in\{0,1\}$ \\
        \textbf{Hidden state:} membrane potential $u_{t}$ \\
        \textbf{Input:} current $[X_{1},\dots, X_{T}]$ \\
        \textbf{Output:} spike $[o_{1},\dots, o_{T}]$
        \begin{algorithmic}[1] 
            \setlength{\baselineskip}{1.1\baselineskip}  
            \FOR{$t = 1$ \TO $T$}
                \STATE $h_{t} = X_{t} + 1/\tau \cdot u_{t-1}$
                \STATE $\rho_{t}=\omega\cdot\rho_{t-1}$ \\
                $\mu_{t}=\mathrm{mean}(h_{t})$, $\sigma_{t}^{2}=\mathrm{variance}(h_{t})$ \\
                $\hat{\mu}=(1-\rho_{t})\cdot\hat{\mu}+\rho_{t}\cdot\mu_{t}$ \\
                $\hat{\sigma^2}=(1-\rho_{t})\cdot\hat{\sigma^2}+\rho_{t}\cdot\sigma^{2}_{t}$ \\
                \STATE $\widetilde{V_{th}} = {(V_{th}-\beta)\cdot\sqrt{\hat{\sigma^{2}}}}/{\gamma} + \hat{\mu}$ \COMMENT{Threshold Modulation}
                \STATE $o_{t} = \mathbf{\Theta}(h_{t} - \widetilde{V_{th}})$
                \STATE $u_{t} = (h_{t}\cdot(1-r)+\mathrm{norm}(h_{t})\cdot r) \cdot (1 - o_{t}) + o_{t} \cdot V_{reset}$
            \ENDFOR
            \IF{$e=1$}
                \STATE compute entropy loss $H$, $\gamma \leftarrow \gamma - \alpha\frac{\partial H}{\partial \gamma}, \space \beta \leftarrow \beta - \alpha\frac{\partial H}{\partial \beta}$
            \ENDIF
        \end{algorithmic}
        \label{algo:tm}
    \end{algorithm}
    \vspace{-23pt}
\end{figure}

Algorithm.~\ref{algo:tm} illustrates the whole internal neuronal dynamics of the LIF neuron with the Threshold Modulation (TM) module. As described earlier, after re-parameterizing the threshold in the pre-trained model, the normalization of each neuron is transformed into an update of the firing threshold, thereby approximating the original neuronal firing dynamics. The hyper-parameters $V_{th}$, $V_{reset}$, and $\tau$ remain consistent with the pre-training phase. The charging and reset phase of the modified LIF neuron remain similar to the original one; however, in the firing phase, the threshold $V_{th}$ is modulated by statistics and affine parameters as in \eqref{eq:Vth}.

The hyper-parameters $\omega$ and $\rho_{0}$ control the momentum-based updates of the statistics, while two flag variables $r$ and $e$ serve as switches for the normalization of the membrane potential of non-firing neurons and the updates of affine parameters $\gamma, \beta$ by entropy minimization: $\mathrm{min}$ $H(\hat{y})= {-\textstyle \sum_{c}^{}p(\hat{y}_{c})\mathrm{log} \space p(\hat{y}_{c})} $, respectively. In practice, whether $\rho_{0}$, $r$, and $e$ are set to $1$ determines the activation of the three components. Specifically, based on whether $e$ is set to $1$ (learnable affine parameters or not), the proposed method is divided into two main variants: \textbf{TM-ENT} and \textbf{TM-NORM}.

\section{Experiments}
\subsection{Datasets}

\textit{1) CIFAR-10/100-C:} We train the source model on CIFAR-10/100 \cite{krizhevskyLearningMultipleLayers2009} image datasets with a training set of $50,000$ and a test set of $10,000$. CIFAR-10-C and CIFAR-100-C \cite{hendrycksBenchmarkingNeuralNetwork2018} apply 15 kinds of common corruptions on the original test set, on which our method will be evaluated.

\textit{2) ImageNet-C:} We also use the ImageNet \cite{russakovskyImageNetLargeScale2015} dataset with a training set of over $1.2$ million and a validation set of $50,000$ images for larger-scale image experiments. ImageNet-C \cite{hendrycksBenchmarkingNeuralNetwork2018} is a corrupted version of the original validation set like CIFAR-10/100-C. We use a fixed subset of $8192$ randomly chosen images in the adaptation experiment.  

\textit{3) SVHN→MNIST/MNIST-M/USPS:} We also evaluate our method's feasibility in simple transfer learning tasks. Following \cite{tangNeuroModulatedHebbianLearning2023a}, we choose SVHN \cite{netzerReadingDigitsNatural2011} as the source domain and transfer the trained model to other digits datasets: MNIST \cite{lecunGradientbasedLearningApplied1998} (with a test set of $10,000$ images), MNIST-M \cite{ganinDomainAdversarialTrainingNeural2016} (with $90,001$ samples of modified MNIST images) and USPS \cite{hullDatabaseHandwrittenText1994} (with a test set of $2,007$ images) respectively.

\subsection{Implementation details}
\label{subsec:implementation}
\paragraph{Model specifications} 
In this work, we adopt spiking Convolutional Neural Networks (CNNs) as the primary model architecture, given their widespread use within the SNN community \cite{wuSpatioTemporalBackpropagationTraining2018,wuDirectTrainingSpiking2019,fangDeepResidualLearning2021,liDifferentiableSpikeRethinking2021,fangIncorporatingLearnableMembrane2021a,cordoneObjectDetectionSpiking2022a,guoMembranePotentialBatch2023a,samadzadehConvolutionalSpikingNeural2023} and as the backbones in TTA research \cite{schneiderImprovingRobustnessCommon2020,wangTentFullyTestTime2020,mirzaNormMustGo2022,niuStableTesttimeAdaptation2022,gongNOTERobustContinual2022,niuEfficientTestTimeModel2022,boudiafParameterfreeOnlineTesttime2022,tangNeuroModulatedHebbianLearning2023a,leeEntropyNotEnough2023}.
When using the spiking CNN models, the first convolution layer can serve as the spike encoder which encodes the continuous input to spike train, mitigating the need for a manually designed one. In image classification models, the final output logits are passed through the Softmax function to compute the predicted class probabilities. While spiking neurons can be employed in the output layer with their firing rates interpreted as logits, we simply use the mean output instead as in \cite{guoMembranePotentialBatch2023a}.

As for model architectures, we use Spiking MobileNet-16, modified VGG-16 with only one fully-connected layer (which we refer to as VGG-16m), ResNet-20, ResNet-19 with an additional fully-connected layer (which we refer to as ResNet-19m) and Wide-ResNet-40-2 for CIAFR-10-C experiments. For CIFAR-100-C experiments, we use spiking ResNet-20 and Wide-ResNet-40-2. For ImageNet-C experiments, we use spiking ResNet-18. The VGG and ResNet models are based on \cite{guoMembranePotentialBatch2023a} and the \textit{spikingjelly} \cite{fangwei123456Fangwei123456Spikingjelly2025} repository. For digit recognition transfer experiment, we use the VGG-like network described in the \textit{pytorch-playground} \cite{chenAaronxichenPytorchplayground2025} repository. All models are first pre-trained on source domain and then prepared for test-time adaptation.

\paragraph{Pre-training hyper-parameters} All models use LIF neuron with $\tau=2$ and  $V_{th}=1$. ResNet and VGG models were trained with the Adam \cite{kingmaAdamMethodStochastic2017} optimizer and an initial learning rate of $0.001$ with a cosine annealing scheduler, the digit model with an initial learning rate of $0.01$ and Wide-ResNet-40-2 models with an initial learning rate of $0.1$. All models are pre-trained with BPTT with sigmoid surrogate gradients ($\alpha=4$) using the SpikingJelly \cite{fangSpikingJellyOpensourceMachine2023a} framework. Data augmentations like random horizontal flip, random crop and cutout are used, and for Wide-ResNet-40-2 we use the AugMix data augmentation \cite{hendrycks*AugMixSimpleData2019}. We trained the models for CIFAR and SVHN for 200 epochs and ImageNet for 320 epochs and then saved the best model on the validation set, serving as a baseline for the tasks. Once a best model is obtained, it is prepared for adaptation as described in \ref{subsec:tm}. 

It is worth noting that due to limited training resources and time, we limited the total number of epochs and are not using other advanced training techniques to improve the model on the source domain. If additional training techniques, or more powerful model architectures were employed to enhance the model's performance to state-of-the-art levels, the final accuracy could be further improved. 

\paragraph{Adaptation experiment setup} The \textbf{adaptation to common corruptions} are conducted like previous TTA research on ANNs \cite{tangNeuroModulatedHebbianLearning2023a,leeEntropyNotEnough2023}. For CIFAR-10/100-C and ImageNet-C, online testing is performed using 15 types of corruptions of the original test set, respectively. The batched data is fed into the network in an online streaming manner. The running accuracy (for online batch input, running accuracy refers to the ratio of correctly classified samples to the total number of samples processed up to the current point) is tracked for each batch and the final running error is reported after the model finished processing all data. For the \textbf{digit recognition transfer task}, the model trained on the SVHN dataset is used for online adaptation on the MNIST, MNIST-M and USPS datasets. The final running error is reported after the model finished processing all the test data. 

\paragraph{Comparison methods}
Since our method assumes that the pre-trained SNN will be deployed on neuromorphic hardware—where batch normalization layers are fused into convolutional layers, model weights are immutable, and input or output modifications are infeasible, leaving only neuron-level adjustments—this represents a novel attempt in the context of TTA. Direct comparisons with the state-of-the-art methods for ANNs are not possible, as they are not applicable in this scenario.
Therefore, we mainly compare our method with the baseline: (i) \textbf{Source}: the pre-trained model is directly tested on the target dataset without adaptation (baseline); (ii) \textbf{TM-NORM}: the pre-trained model adapts to the corrupted data in the target domain with Threshold Modulation and the affine parameters are frozen; (iii) \textbf{TM-ENT}: the pre-trained model adapts to the corrupted data in the target domain with Threshold Modulation and the affine parameters are updated by entropy minimization. Batch size is set to $64$ to represent scenarios with relatively abundant resources. For CIFAR-10-C, batch size of $1$ is also included to represent highly resource-constrained single-image online adaptation settings. The initial momentum $\rho_{0}$ is set to $1.0$ in CIFAR and ImageNet tasks, and $0.9$ in the digit recognition transfer task. The momentum decay $\omega$ is set to $0.94$. We set $r=0$ in the adaptation to common corruptions task and $r=1$ in the digits recognition transfer task. A fixed random seed is selected for all experiment. For trails using TM-ENT, we use Adam \cite{kingmaAdamMethodStochastic2017} optimizer with a learning rate of $0.00025$ for $bs=64$ and $0.00025/16$ for $bs=1$. 

Additionaly, for CIFAR-10-C and spiking ResNet-20, we also include comparisons with two classical methods in TTA: \textbf{NORM} \cite{schneiderImprovingRobustnessCommon2020} (with $N\!=\!0$ to align with TM-NORM with $\rho_{0}\!=\!1$) and \textbf{TENT} \cite{wangTentFullyTestTime2020}. These two methods correspond to BN calibration and entropy minimization, and we apply them directly on a pre-trained SNN without MPBN. It must be emphasized that these methods cannot be implemented in the hardware-friendly scenarios described in this paper, and thus, they are provided for reference only and not directly compared with our method in terms of performance.

\subsection{Results}
\label{subsec:results}

\begin{table*}[t]
    \setlength{\abovecaptionskip}{0pt}
    \caption{Top-1 Classification Error (\%) for each corruption in \textbf{CIFAR-10-C} at the highest severity (Level \textbf{5}). \\  The 15 columns represent 15 different types of corruption. The lowest error rates are highlighted in bold.}
    \centering
    \begin{tabular}{@{\hskip 1pt}c@{\hskip 1pt}|@{\hskip 3pt}c@{\hskip 3pt}|p{14pt}p{14pt}p{14pt}p{14pt}p{14pt}p{14pt}p{14pt}p{14pt}p{14pt}p{14pt}p{14pt}p{14pt}p{14pt}p{14pt}p{14pt}|r}
        \toprule
        Network & Method & gaus & shot & impul & defcs & gls & mtn & zm & snw & frst & fg & brt & cnt & els & px & jpg & Avg.\\
        \midrule
        \multirow{5}*{\parbox{44pt}{\centering \vspace{15pt} ResNet \\ 20}} & Source & 72.5 & 66.9 & 71.3 & 48.1 & 56.8 & 43.4 & 38.0 & 28.6 & 37.7 & 33.9 & \textbf{11.3} & 76.4 & 29.3 & 59.5 & 27.0 & 46.7 \\
        & TM-NORM & 34.7 & 31.9 & 40.0 & 17.2 & 39.1 & 19.1 & \textbf{16.3} & 25.4 & 24.5 & 20.2 & 11.7 & 18.9 & 26.2 & 24.7 & 27.8 & 25.2 \\
        & TM-ENT & \textbf{34.5} & \textbf{31.1} & \textbf{39.7} & \textbf{17.0} & \textbf{38.3} & \textbf{18.7} & 16.4 & \textbf{25.1} & \textbf{24.0} & \textbf{20.0} & 11.4 & \textbf{18.7} & \textbf{25.6} & \textbf{24.3} & \textbf{26.9} & \textbf{24.8} \\
        \noalign{\smallskip}
        \cline{2-18}
        \noalign{\smallskip}
        & Source & 73.0 & 67.4 & 69.3 & 49.7 & 61.7 & 43.9 & 40.2 & 31.2 & 42.2 & 34.2 & 11.8 & 78.0 & 30.6 & 59.0 & 28.4 & 48.0 \\
        & NORM$^{\mathrm{a}}$ & 32.4 & 28.9 & 38.7 & 15.8 & 37.2 & 17.4 & 15.1 & 22.5 & 21.3 & 18.1 & 10.6 & 16.2 & 25.0 & 24.0 & 25.7 & 23.7 \\
        & TENT$^{\mathrm{a}}$ & 32.1 & 28.9 & 38.7 & 16.0 & 37.0 & 17.4 & 15.1 & 22.4 & 21.4 & 18.0 & 10.6 & 15.6 & 25.1 & 23.6 & 25.5 & 23.2 \\
        \noalign{\smallskip}
        \hline
        \noalign{\smallskip}
        \multirow{3}*{\parbox{44pt}{\centering ResNet \\ 19m}} & Source & 68.7 & 63.0 & 67.2 & 50.0 & 58.6 & 47.6 & 42.6 & 31.5 & 38.7 & 37.5 & 13.0 & 76.2 & 32.1 & 60.5 & 28.3 & 47.7 \\
        & TM-NORM & \textbf{36.0} & \textbf{33.7} & \textbf{42.1} & \textbf{17.5} & \textbf{40.5} & \textbf{19.9} & \textbf{16.3} & \textbf{25.0} & \textbf{25.1} & \textbf{20.4} & \textbf{11.6} & \textbf{19.6} & \textbf{27.5} & \textbf{25.3} & \textbf{27.7} & \textbf{25.9} \\
        & TM-ENT & 36.4 & 34.3 & 43.0 & 17.9 & 41.0 & 20.0 & 16.7 & 25.5 & 25.2 & 20.7 & 11.9 & 20.0 & 27.8 & 26.0 & 28.1 & 26.3 \\
        \noalign{\smallskip}
        \hline
        \noalign{\smallskip}
        \multirow{3}*{\parbox{44pt}{\centering VGG \\ 16m}} & Source & 64.0 & 57.4 & 70.2 & 52.0 & 53.4 & 45.0 & 44.4 & 31.4 & 40.4 & 42.0 & 14.7 & 79.0 & 26.4 & 50.9 & \textbf{22.8} & 46.3 \\
        & TM-NORM & 32.4 & 31.1 & 37.5 & \textbf{18.1} & 37.0 & \textbf{21.0} & \textbf{18.4} & \textbf{26.9} & \textbf{25.6} & \textbf{24.7} & \textbf{13.5} & \textbf{25.4} & \textbf{23.0} & 24.8 & 23.1 & \textbf{25.5} \\
        & TM-ENT & \textbf{32.1} & \textbf{31.0} & \textbf{37.3} & 18.2 & \textbf{36.7} & 21.1 & 18.6 & 27.1 & 25.7 & 24.8 & 13.7 & 25.6 & 23.5 & \textbf{24.5} & 23.3 & 25.6 \\
        \noalign{\smallskip}
        \hline
        \noalign{\smallskip}
        \multirow{3}*{\parbox{44pt}{\centering MobileNet \\ 16}} & Source & 74.2 & 71.0 & 71.6 & 61.7 & 53.5 & 49.4 & 56.1 & 41.3 & 54.5 & 54.8 & 20.9 & 78.3 & 29.5 & 59.7 & \textbf{28.2} & 53.6 \\
        & TM-NORM & \textbf{47.0} & \textbf{45.5} & 44.7 & \textbf{24.7} & 40.2 & 27.8 & 24.7 & 34.7 & \textbf{35.4} & 37.8 & 21.5 & \textbf{37.3} & 30.0 & \textbf{32.5} & 29.9 & 34.3 \\
        & TM-ENT & 47.1 & 46.1 & \textbf{44.2} & \textbf{24.7} & \textbf{40.0} & \textbf{27.7} & \textbf{24.0} & \textbf{34.2} & \textbf{35.4} & \textbf{37.7} & \textbf{21.4} & \textbf{37.3} & \textbf{28.9} & 32.8 & 30.3 & \textbf{34.1} \\
        \noalign{\smallskip}
        \hline
        \noalign{\smallskip}
        \multirow{5}*{\parbox{44pt}{\centering WideResNet \\ 40-2}} & Source & 49.7 & 44.7 & 43.7 & 17.5 & 43.7 & 22.2 & 18.0 & 27.9 & 34.9 & 31.6 & 14.3 & 40.9 & 26.7 & 39.2 & 24.7 & 32.0 \\
        & TM-NORM & 26.1 & 23.8 & 23.0 & 14.5 & \textbf{29.1} & 17.2 & 14.9 & 20.4 & \textbf{18.8} & 22.6 & 11.9 & 15.2 & 21.3 & \textbf{21.5} & 22.7 & 20.2 \\
        & TM-ENT & \textbf{25.4} & \textbf{23.5} & \textbf{22.7} & \textbf{14.0} & 29.2 & \textbf{16.9} & \textbf{14.4} & \textbf{19.9} & 19.0 & \textbf{21.8} & \textbf{11.3} & \textbf{15.1} & \textbf{20.8} & 21.7 & \textbf{22.1} & \textbf{19.9} \\
        & TM-NORM$^{\mathrm{b}}$ & 28.1 & 25.8 & 24.8 & 17.5 & 33.1 & 21.0 & 18.0 & 24.5 & 22.7 & 24.8 & 14.5 & 26.0 & 24.5 & 26.0 & 26.8 & 23.9 \\
        & TM-ENT$^{\mathrm{b}}$ & 27.7 & 25.9 & 24.6 & 17.3 & 33.3 & 20.7 & 17.4 & 24.5 & 22.3 & 24.7 & 14.5 & 26.1 & 24.1 & 25.4 & 26.2 & 23.6 \\
        \bottomrule
        \multicolumn{12}{l}{$^{\mathrm{a}}$ model pre-trained without TM; cannot be applied directly in the discussed on-chip scenario.}\\
        \multicolumn{4}{l}{$^{\mathrm{b}}$ $batchsize=1$.}
    \end{tabular}
    \label{tab:cifar10c}
    \vspace{-12pt}
\end{table*}

\begin{table*}[t]
    \setlength{\abovecaptionskip}{0pt}
    \caption{Top-1 Classification Error (\%) for each corruption in \textbf{CIFAR-100-C} at the highest severity (Level \textbf{5}).}
    \centering
    \begin{tabular}{@{\hskip 1pt}c@{\hskip 1pt}|@{\hskip 5pt}c@{\hskip 5pt}|p{14pt}p{14pt}p{14pt}p{14pt}p{14pt}p{14pt}p{14pt}p{14pt}p{14pt}p{14pt}p{14pt}p{14pt}p{14pt}p{14pt}p{14pt}|r}
        \toprule
        Network & Method & gaus & shot & impul & defcs & gls & mtn & zm & snw & frst & fg & brt & cnt & els & px & jpg & Avg.\\
        \midrule
        \multirow{3}*{\parbox{44pt}{\centering ResNet \\ 20}} & Source & 88.3 & 86.1 & 89.6 & 72.8 & 79.8 & 66.3 & 66.8 & 59.0 & 67.9 & 70.6 & 43.9 & 90.5 & 57.3 & 78.8 & 57.4 & 71.7 \\
        & TM-NORM & 62.3 & 60.7 & 66.0 & 41.4 & 60.8 & 44.9 & 41.4 & 53.0 & 53.5 & 52.3 & \textbf{38.3} & 46.1 & 50.1 & 47.7 & 55.2 & 51.6 \\
        & TM-ENT & \textbf{61.8} & \textbf{60.4} & \textbf{65.8} & \textbf{41.1} & \textbf{60.4} & \textbf{44.7} & \textbf{40.8} & \textbf{52.1} & \textbf{53.1} & \textbf{51.4} & 38.5 & \textbf{45.8} & \textbf{49.5} & \textbf{47.1} & \textbf{54.6} & \textbf{51.1} \\
        \noalign{\smallskip}
        \hline
        \noalign{\smallskip}
        \multirow{3}*{\parbox{44pt}{\centering Wide \\ ResNet \\ 40-2}} & Source & 87.8 & 85.9 & 81.6 & 52.6 & 78.6 & 57.9 & 53.0 & 65.8 & 76.7 & 73.3 & 54.4 & 77.2 & 65.0 & 70.4 & 61.6 & 69.5 \\
        & TM-NORM & 57.2 & 55.3 & 49.3 & 41.6 & 55.9 & 44.7 & 42.6 & 48.9 & 48.8 & 53.3 & 39.3 & 44.4 & 47.6 & 49.1 & 52.0 & 48.7 \\
        & TM-ENT & \textbf{56.8} & \textbf{54.9} & \textbf{49.0} & \textbf{41.4} & \textbf{55.0} & \textbf{44.0} & \textbf{41.8}& \textbf{48.7} & \textbf{48.2} & \textbf{53.0} & \textbf{38.5} & \textbf{44.3} & \textbf{46.9} & \textbf{47.7} & \textbf{51.7} & \textbf{48.1} \\
        \bottomrule
    \end{tabular}
    \label{tab:cifar100c}
    \vspace{-2pt}
\end{table*}
    
\begin{table*}[t]
    \setlength{\abovecaptionskip}{0pt}
    \caption{Top-1 Classification Error (\%) for each corruption in \textbf{ImageNet-C} at severity level \textbf{1}.}
    \centering
    \begin{tabular}{@{\hskip 1pt}c@{\hskip 1pt}|@{\hskip 5pt}c@{\hskip 5pt}|p{14pt}p{14pt}p{14pt}p{14pt}p{14pt}p{14pt}p{14pt}p{14pt}p{14pt}p{14pt}p{14pt}p{14pt}p{14pt}p{14pt}p{14pt}|r}
        \toprule
        Network & Method & gaus & shot & impul & defcs & gls & mtn & zm & snw & frst & fg & brt & cnt & els & px & jpg & Avg.\\
        \midrule
        \multirow{3}*{\parbox{44pt}{\centering ResNet\\18}} & Source & 70.1 & 72.8 & 78.8 & 81.3 & 77.9 & 73.3 & 81.2 & 76.9 & 72.5 & 86.5 & 54.4 & 82.1 & 65.5 & 61.8 & 62.4 & 73.2 \\
        & TM-NORM & 61.3 & 63.2 & \textbf{71.7} & 64.0 & \textbf{61.5} & 56.3 & \textbf{62.3} & 64.3 & 59.6 & 57.0 & \textbf{46.4} & \textbf{53.6} & \textbf{50.7} & 51.7 & 52.7 & 58.4 \\
        & TM-ENT & \textbf{61.2} & \textbf{62.9} & 72.0 & \textbf{63.4} & 61.8 & \textbf{55.3} & 62.4 & \textbf{63.4} & \textbf{59.0} & \textbf{56.6} & 46.6 & 53.9 & 51.1 & \textbf{51.4} & \textbf{52.3} & \textbf{58.2} \\
        \bottomrule
    \end{tabular}
    \label{tab:imagenetc}
    \vspace{-12pt}
\end{table*}

\begin{table}[t]
    \setlength{\abovecaptionskip}{0pt}
    \caption{Top-1 Classification Error (\%) \\ for digit recognition transfer from SVHN.}
    \centering
    \begin{tabular}{l|ccc|r}
        \toprule
        Method & MNIST & MNIST-M & USPS & Avg.\\			
        \midrule
        Source & 53.46 & \textbf{53.09} & \textbf{26.06} & 44.20 \\
        TM-NORM & 29.66 & 54.24 & 29.80 & 37.90 \\
        TM-ENT & \textbf{28.47} & 53.33 & 28.75 & \textbf{36.85} \\
        \bottomrule
    \end{tabular}
    \label{tab:svhn}
    \vspace{-12pt}
\end{table}

\textbf{Adaptation to common corruptions.}
In the corruption benchmark, The classification errors are reported in Tab.~\ref{tab:cifar10c}, ~\ref{tab:cifar100c} and Tab.~\ref{tab:imagenetc}. Tab.~\ref{tab:cifar10c} compares the classification error on CIFAR-10-C during adaptation. We conducted experiments using three model backbones: ResNet-20, ResNet-19m, VGG-16m and Wide-ResNet-40-2, demonstrating the effectiveness of the proposed test-time adaptation method TM-TTA. Notably, the results for Wide-ResNet-40-2 highlight that targeted optimizations applied to the training set, combined with our method, can further enhance model robustness. The extreme case with a batch size of $1$ demonstrates that, although our method relies on online data statistics, it can still deliver usable performance in single-image online streaming setting. Comparison with NORM and TENT reveals that our method achieves performance comparable to the corresponding methods for ANNs while being neuromorphic chip-friendly. Tab.~\ref{tab:cifar100c} presents the classification error during test-time adaptation on CIFAR-100-C, showing similar trends to those observed on CIFAR-10-C. For the more challenging ImageNet-C classification task, the results in Tab.~\ref{tab:imagenetc} also demonstrate the performance of the proposed method in more complicated datasets. 

\begin{figure*}[t]
    \centering
    \begin{minipage}{0.04\linewidth}
        \centering
        \rotatebox{90}{\fontsize{8pt}{10pt}\selectfont Channel}
    \end{minipage}%
    \begin{minipage}{0.24\linewidth}
        \centering
        \includegraphics[width=\linewidth]{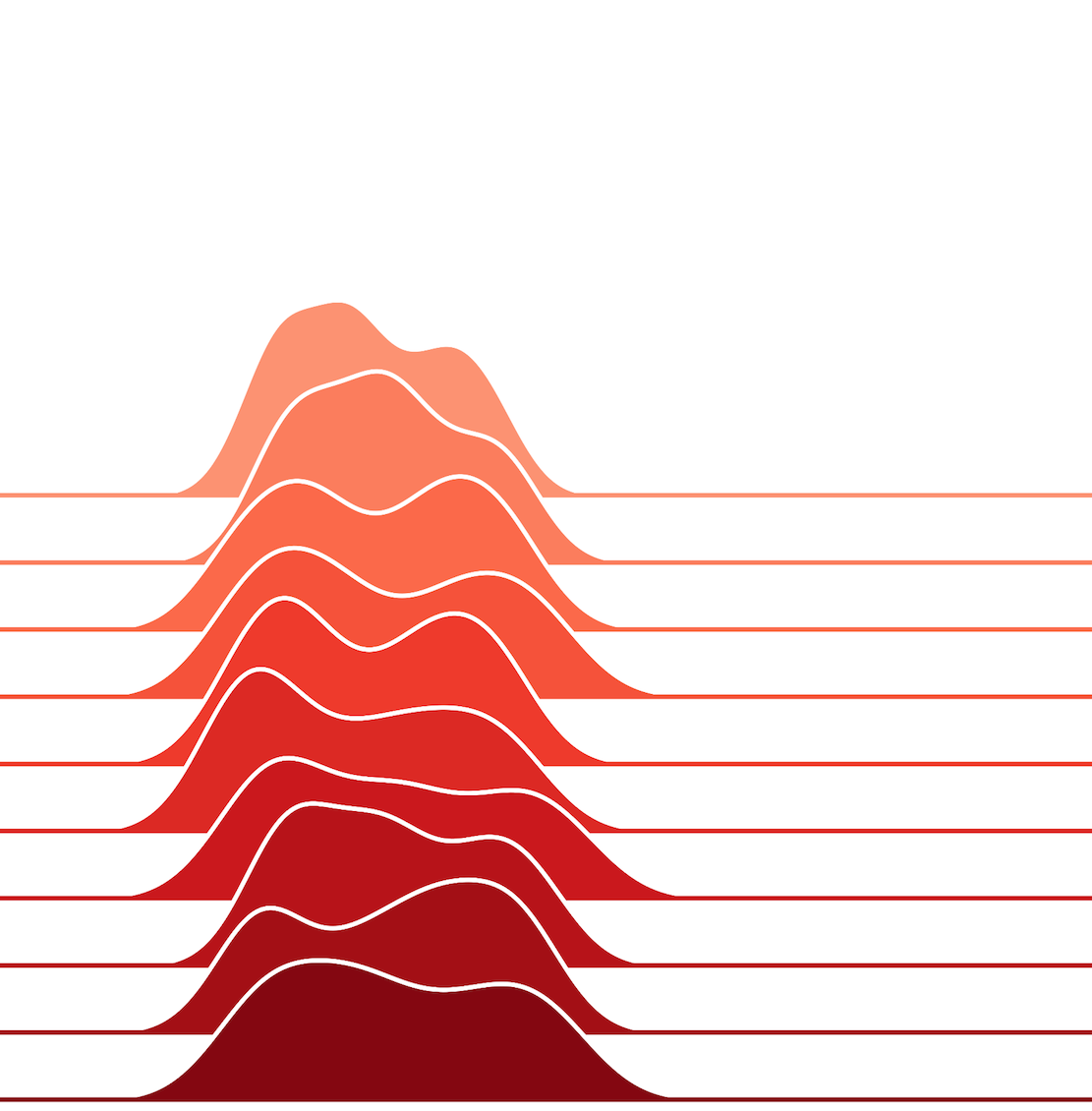}
        \label{subfig:channel_clean}
    \end{minipage}%
    \begin{minipage}{0.24\linewidth}
        \centering
        \includegraphics[width=\linewidth]{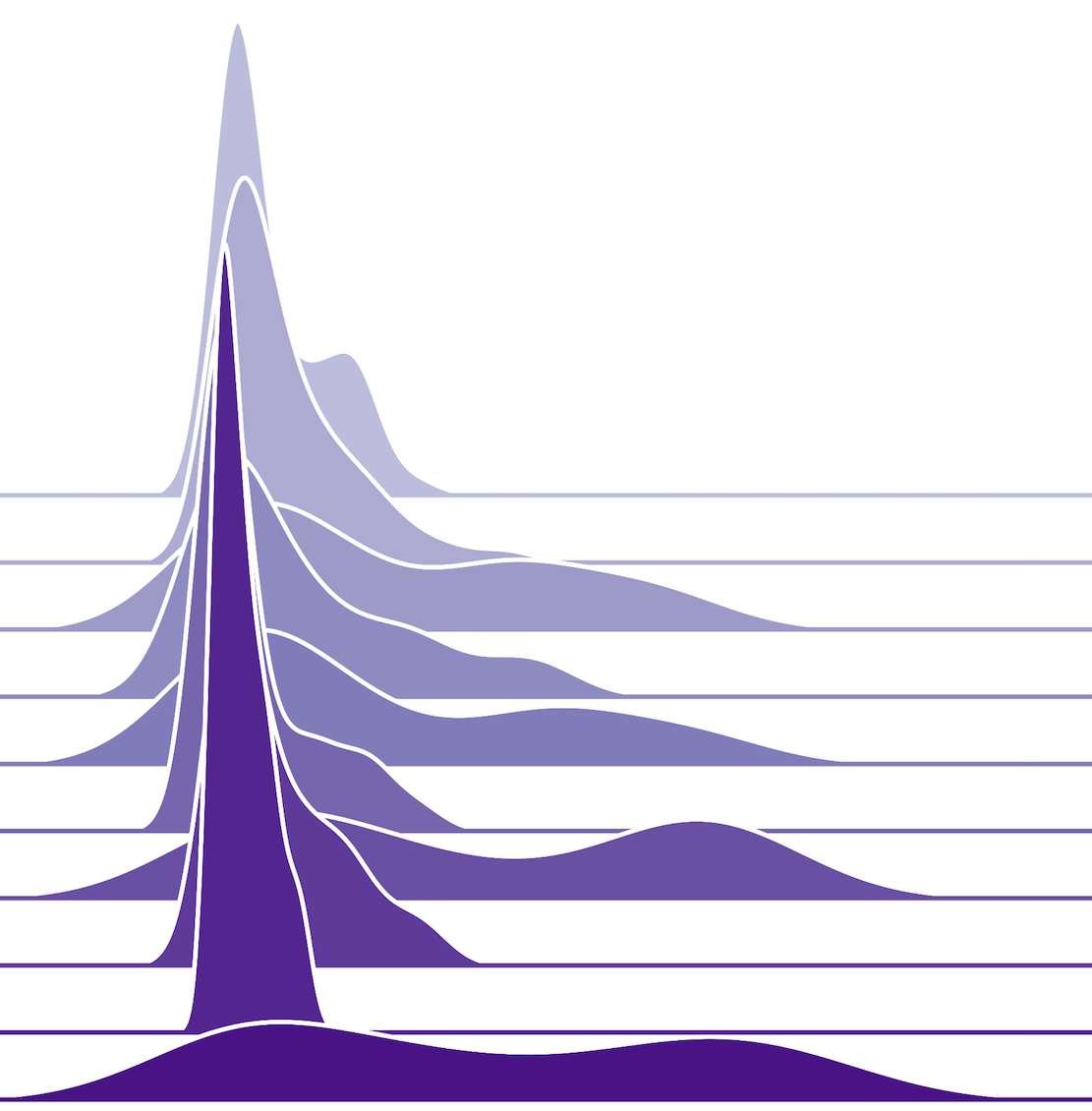}
        \label{subfig:channel_trp}
    \end{minipage}%
    \begin{minipage}{0.24\linewidth}
        \centering
        \includegraphics[width=\linewidth]{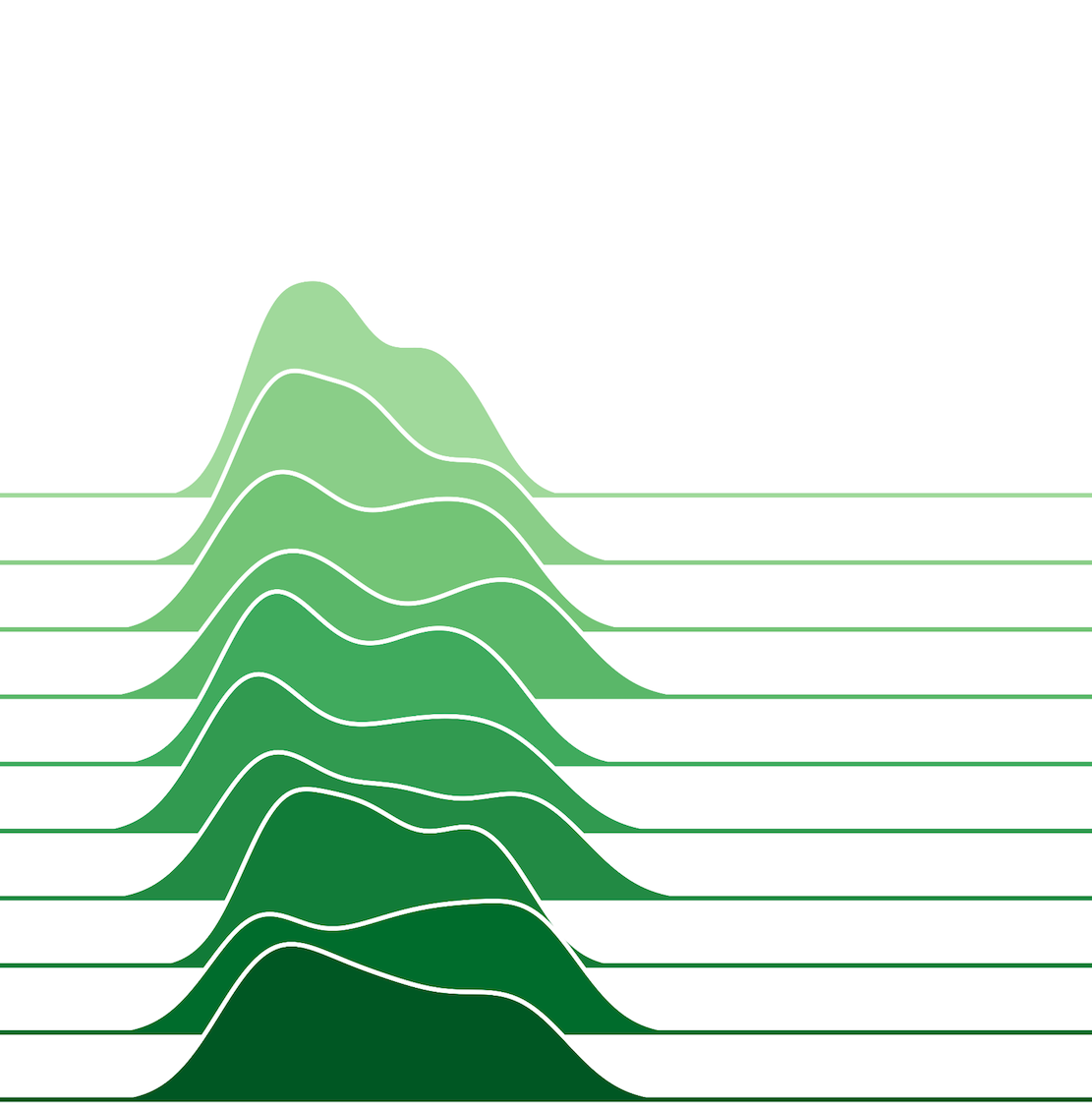}
        \label{subfig:channel_norm}
    \end{minipage}%
    \begin{minipage}{0.24\linewidth}
        \centering
        \includegraphics[width=\linewidth]{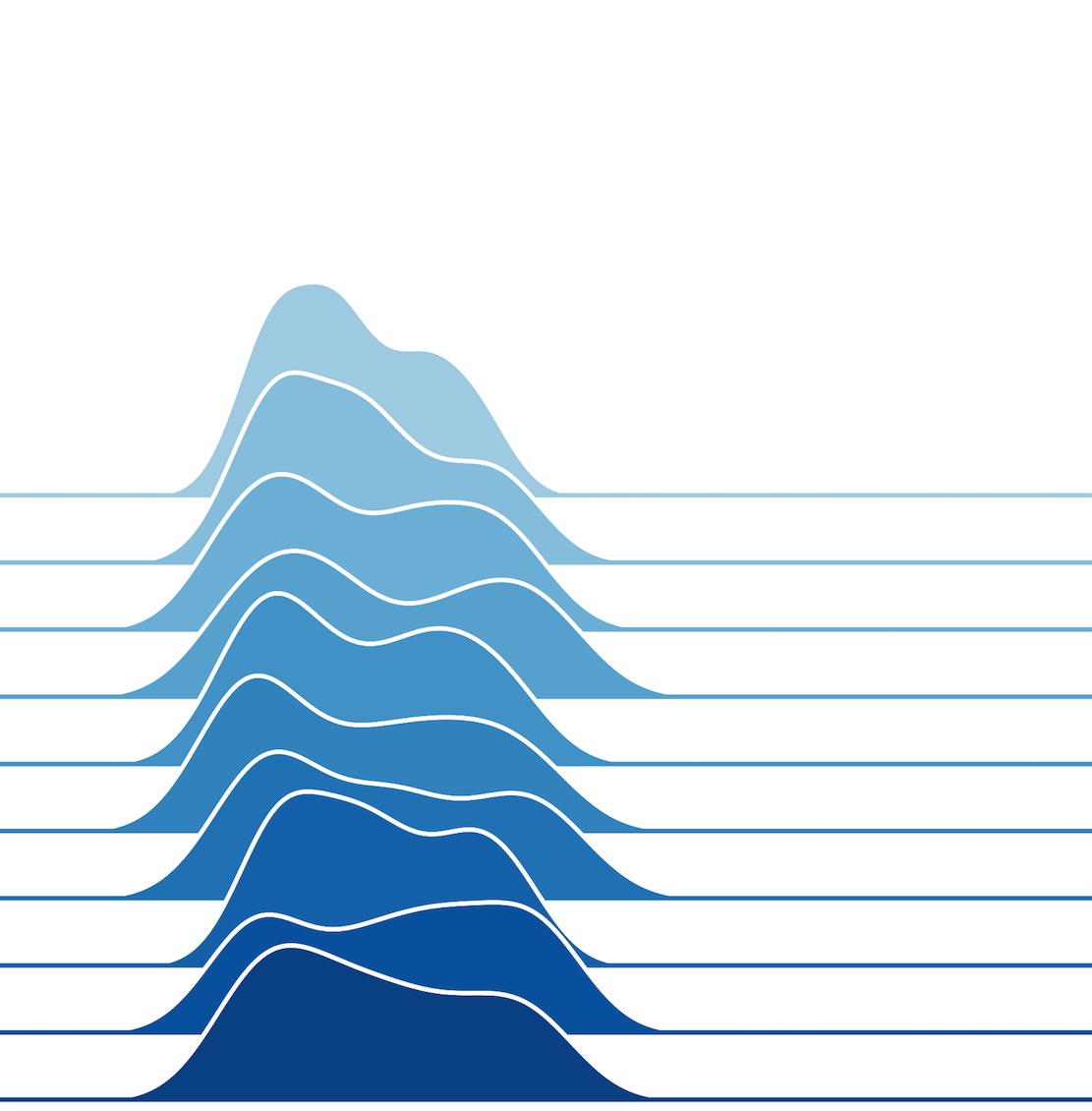}
        \label{subfig:channel_source}
    \end{minipage}%
    \vspace{-\baselineskip}
    \begin{minipage}{0.04\linewidth}
        \centering
        \rotatebox{90}{\fontsize{8pt}{10pt}\selectfont Layer 1}
    \end{minipage}%
    \begin{minipage}{0.24\linewidth}
        \centering
        \includegraphics[width=\linewidth]{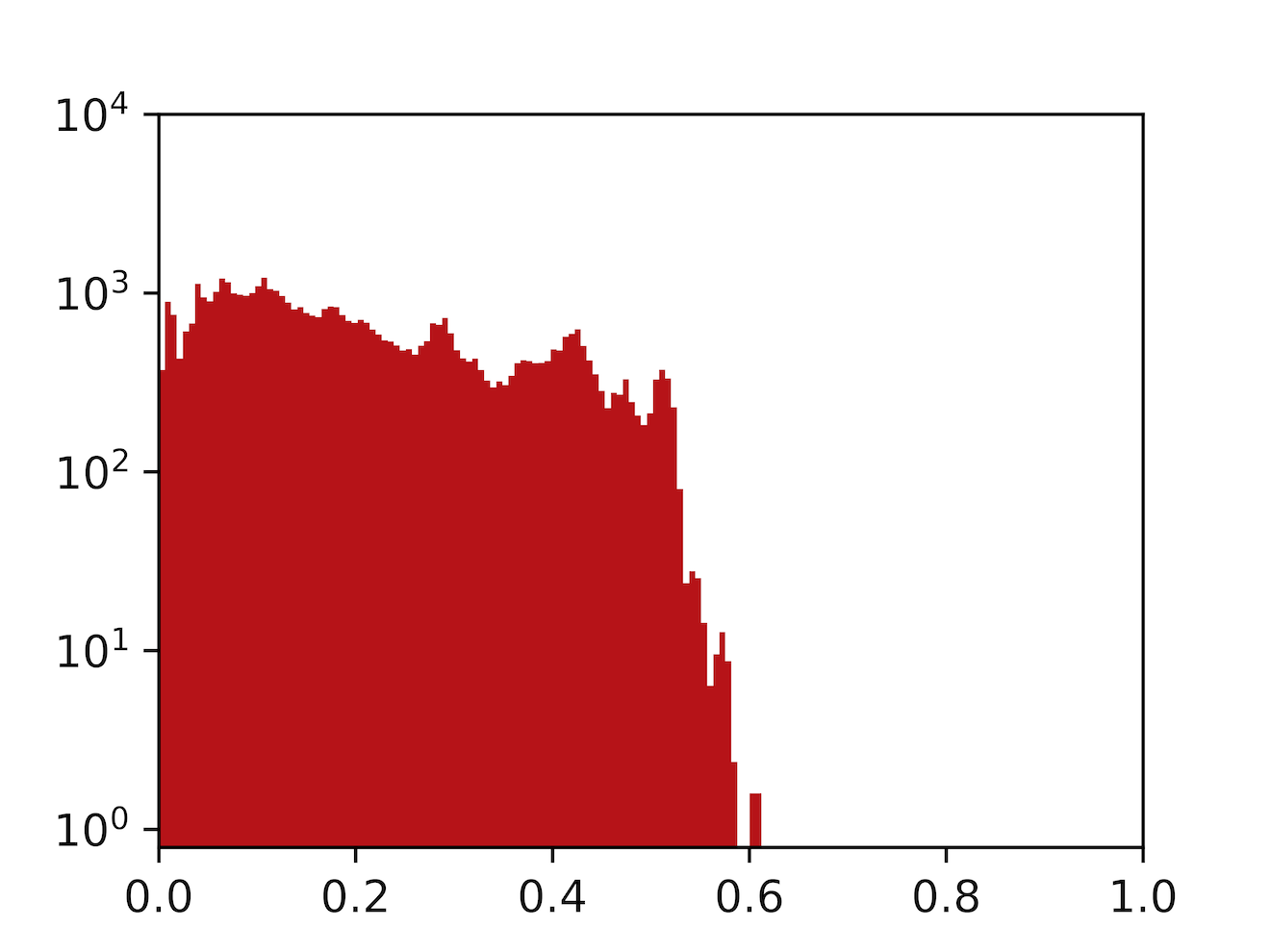}
        \label{subfig:clean_l1}
    \end{minipage}%
    \begin{minipage}{0.24\linewidth}
        \centering
        \includegraphics[width=\linewidth]{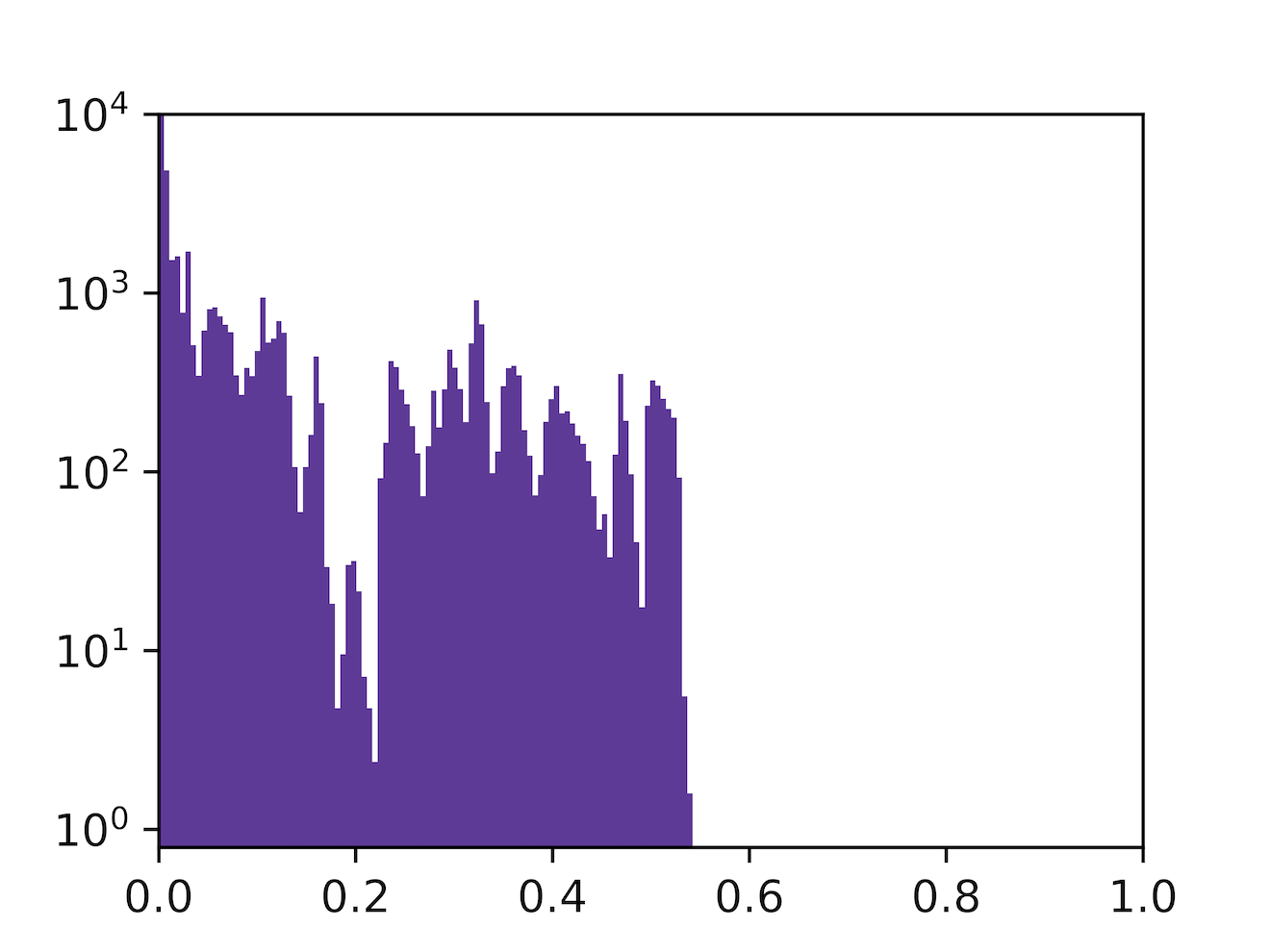}
        \label{subfig:source_l1}
    \end{minipage}%
    \begin{minipage}{0.24\linewidth}
        \centering
        \includegraphics[width=\linewidth]{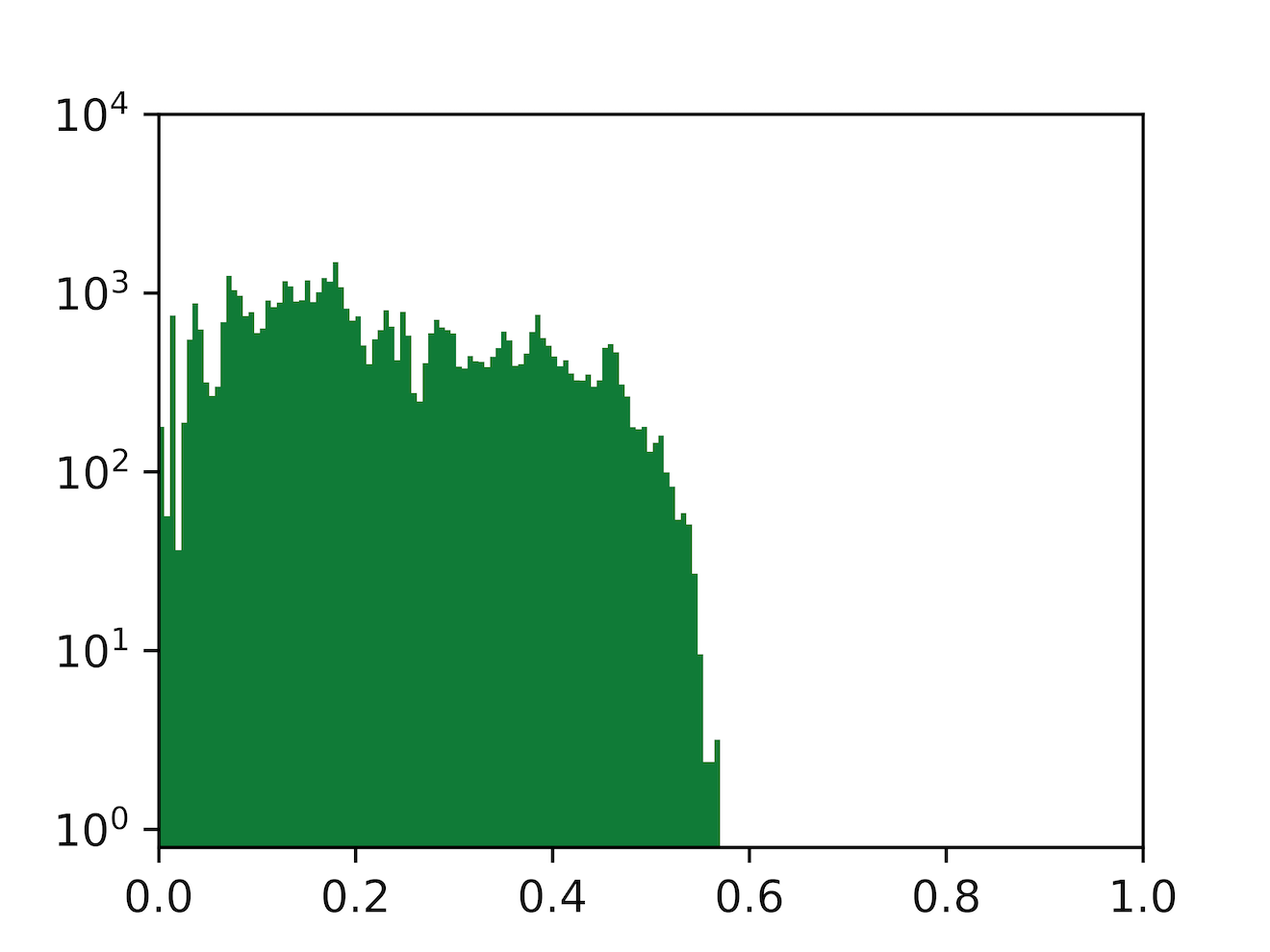}
        \label{subfig:norm_l1}
    \end{minipage}%
    \begin{minipage}{0.24\linewidth}
        \centering
        \includegraphics[width=\linewidth]{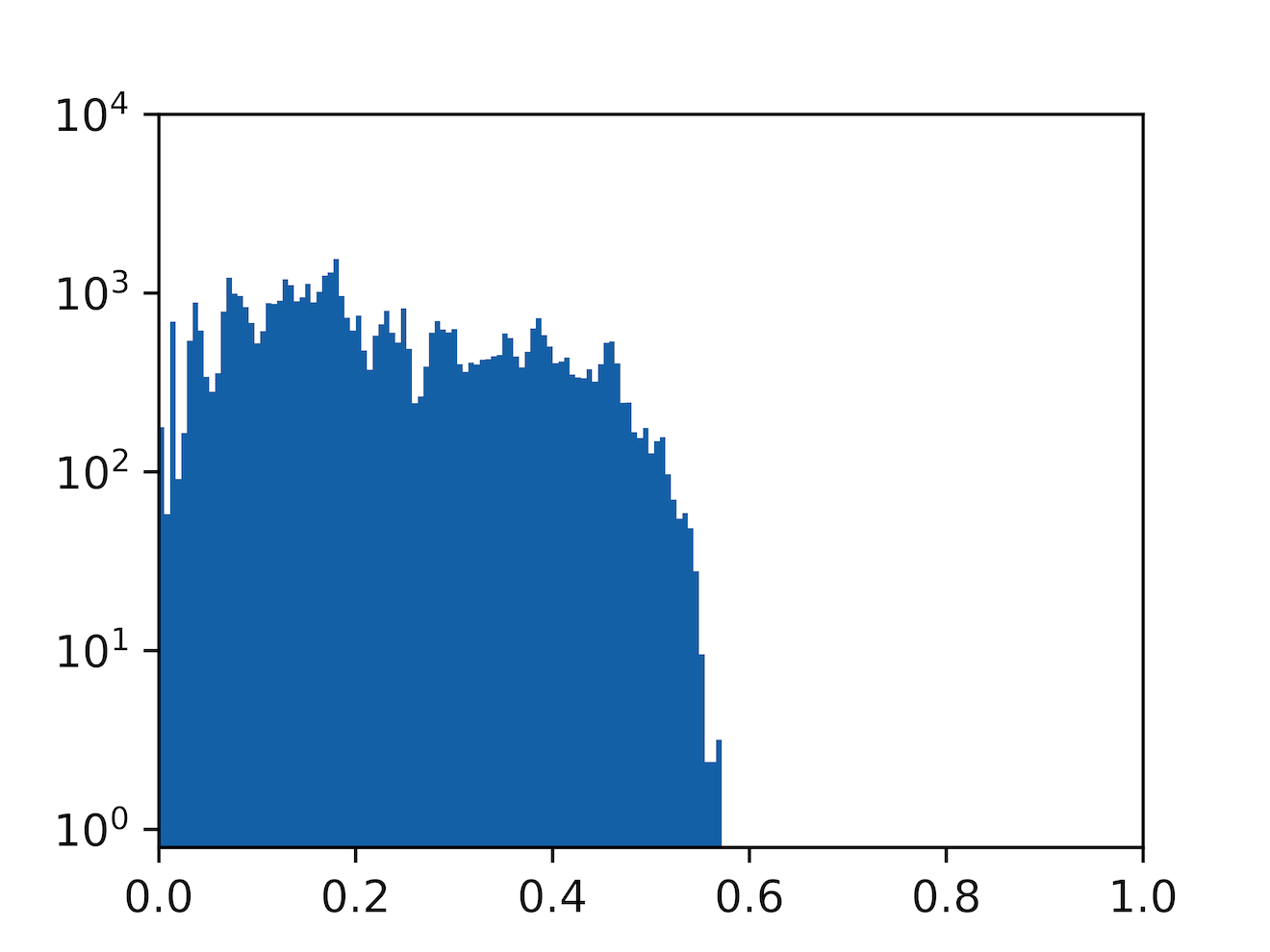}
        \label{subfig:trp_l1}
    \end{minipage}%
    \vspace{-\baselineskip}
    \begin{minipage}{0.04\linewidth}
        \centering
        \rotatebox{90}{\fontsize{8pt}{10pt}\selectfont Layer 6}
    \end{minipage}%
    \begin{minipage}{0.24\linewidth}
        \centering
        \includegraphics[width=\linewidth]{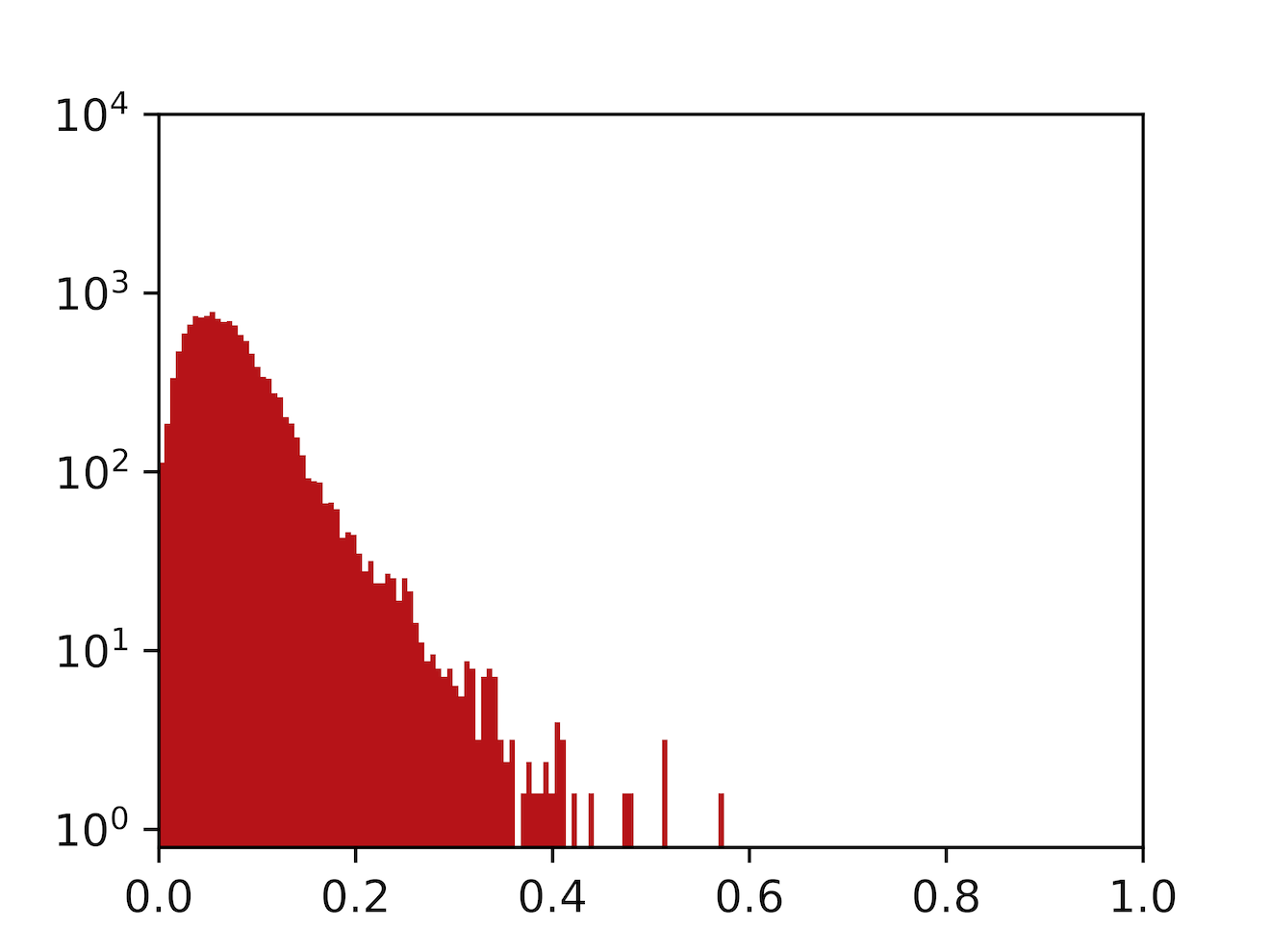}
        \label{subfig:clean_l5}
    \end{minipage}%
    \begin{minipage}{0.24\linewidth}
        \centering
        \includegraphics[width=\linewidth]{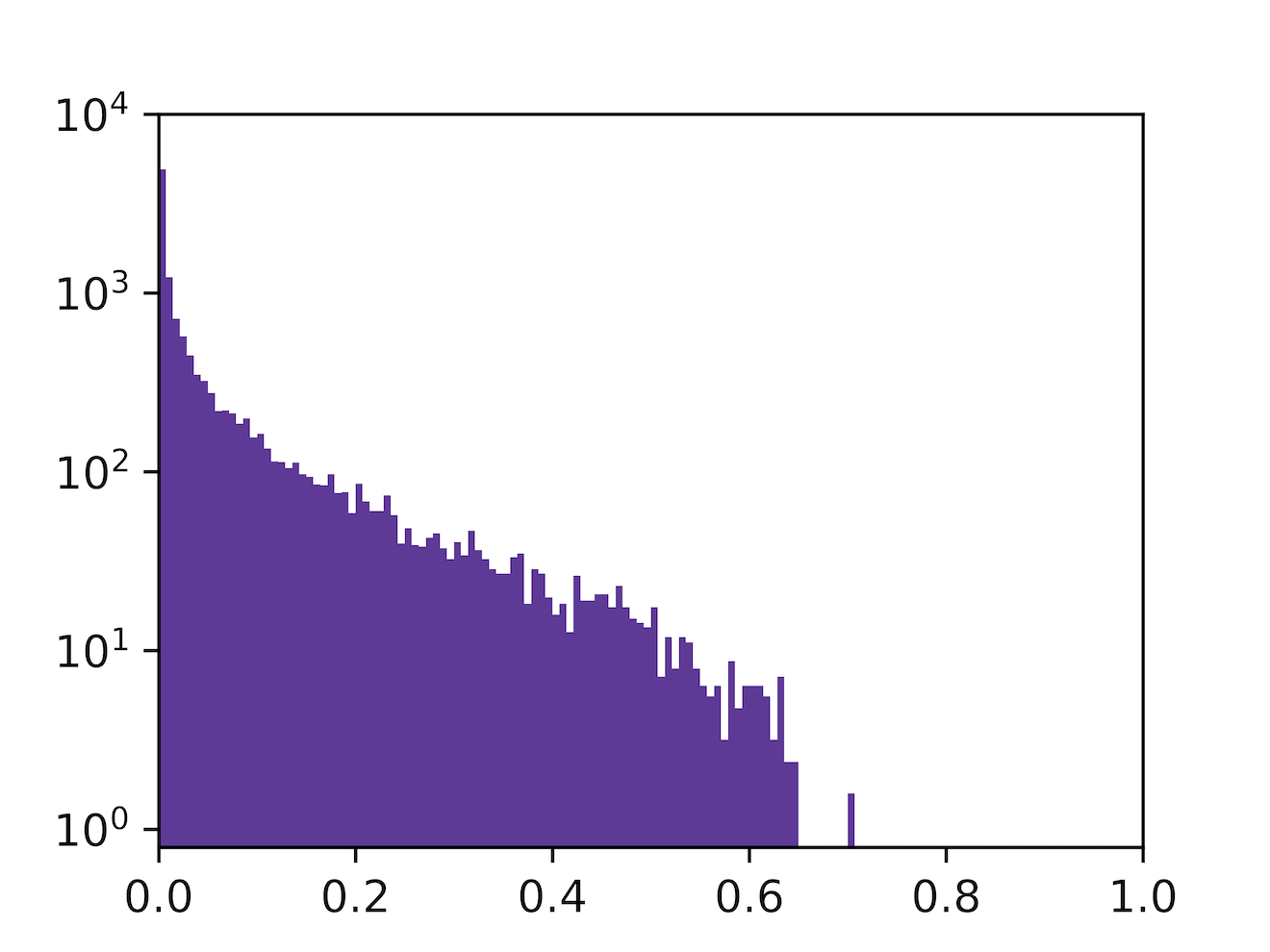}
        \label{subfig:source_l5}
    \end{minipage}%
    \begin{minipage}{0.24\linewidth}
        \centering
        \includegraphics[width=\linewidth]{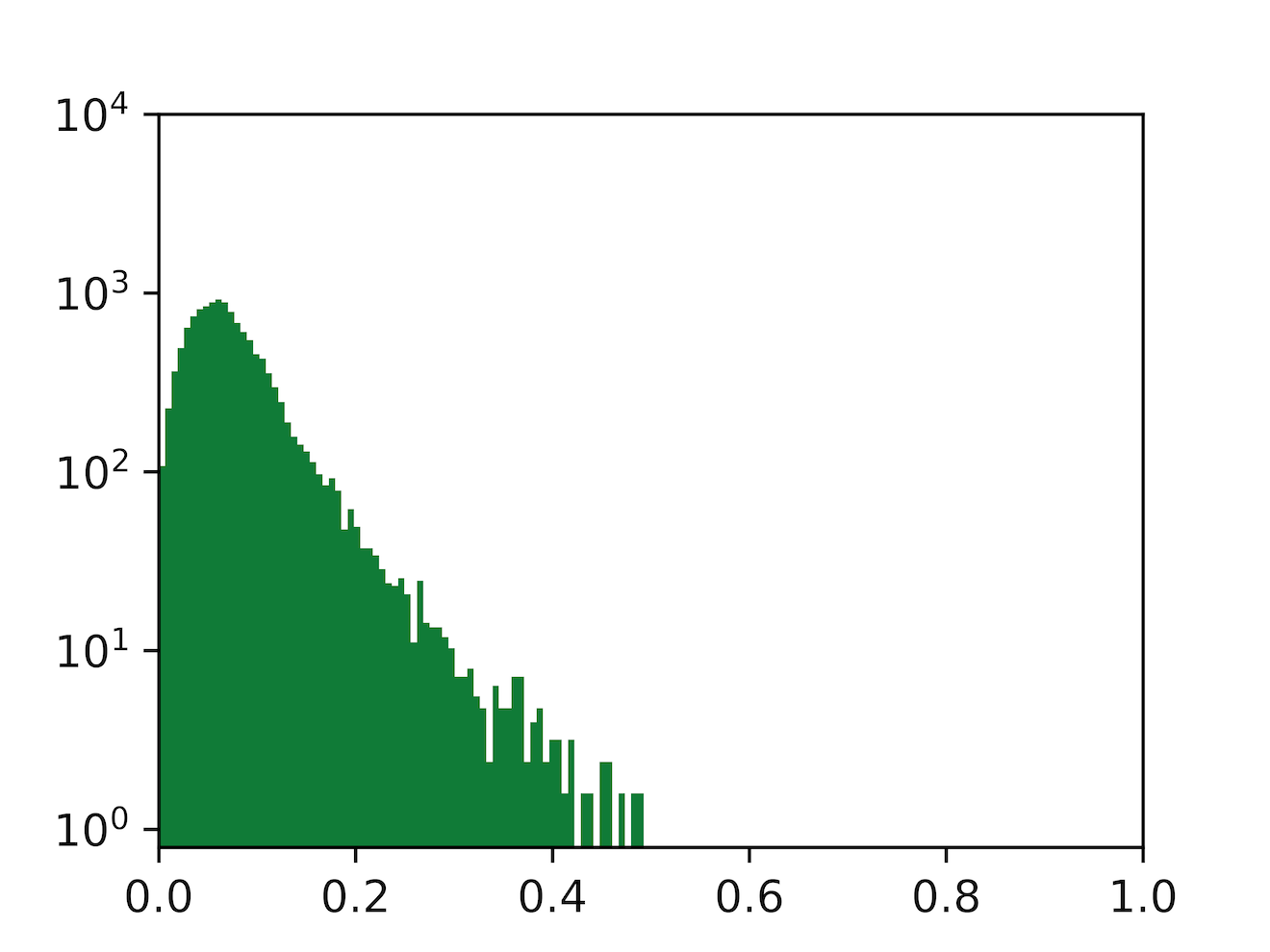}
        \label{subfig:norm_l5}
    \end{minipage}%
    \begin{minipage}{0.24\linewidth}
        \centering
        \includegraphics[width=\linewidth]{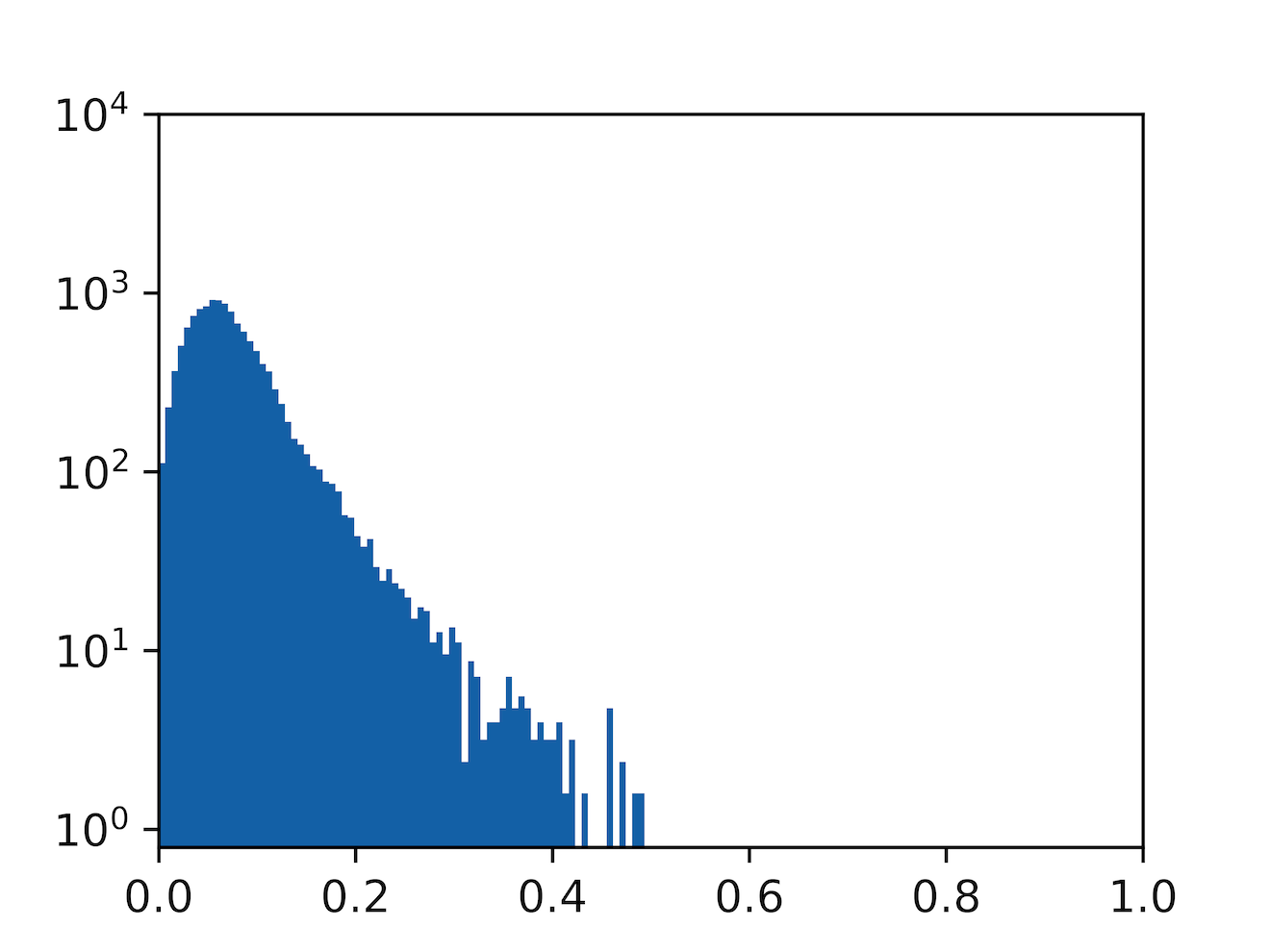}
        \label{subfig:trp_l5}
    \end{minipage}%
    \vspace{-\baselineskip}
    \begin{minipage}{0.04\linewidth}
        \centering
        \phantom{}
    \end{minipage}%
    \begin{minipage}{0.24\linewidth}
        \centering
        {\fontsize{8pt}{10pt}\selectfont
        Clean}
    \end{minipage}%
    \begin{minipage}{0.24\linewidth}
        \centering
        {\fontsize{8pt}{10pt}\selectfont
        Source}
    \end{minipage}%
    \begin{minipage}{0.24\linewidth}
        \centering
        {\fontsize{8pt}{10pt}\selectfont
        TM-NORM}
    \end{minipage}%
    \begin{minipage}{0.24\linewidth}
        \centering
        {\fontsize{8pt}{10pt}\selectfont
        TM-ENT}
    \end{minipage}%
    \vspace{-4pt}
    \caption{\textbf{Firing rate visualization on CIFAR-10-C with Contrast.} \textbf{Upper}: Density Estimation of firing rates of the last convolutional layer in Spiking ResNet-20. \textbf{Lower}: Histogram of firing rates of the two layers in Spiking VGG-16m. The firing rate distribution was shifted away by corrupted data ('Source'); TM-NORM and TM-ENT bring it closer to the clean data.}
    \label{fig:firingrate}
    \vspace{-10pt}
\end{figure*}

Fig.~\ref{fig:firingrate} presents a visualization of the firing rates. The distribution of firing rates indicates that dataset distribution shift leads to significant changes in the firing rate distribution, thereby affecting the performance of the network. The proposed TM-NORM and TM-ENT methods can effectively correct the firing rate shift, bringing it closer to that of the original dataset, thus mitigating the performance degradation caused by the shift in dataset distribution. 

\textbf{Digit recognition transfer task.}
Following \cite{wangTentFullyTestTime2020, tangNeuroModulatedHebbianLearning2023a}, we report experimental results for digit recognition transfer task from SVHN to MNIST, MNIST-M and USPS datasets in Tab.~\ref{tab:svhn}. It can be observed that the proposed method significantly reduces the classification error on the MNIST dataset, while a slight increase in the final error rate is noted on the MNIST-M and USPS datasets. Specifically, on these two datasets, the classification error rate decreases substantially during the initial batches. However, as the number of input batches increases, the error rate experiences a slight rise before stabilizing. This phenomenon is also observed in \cite{tangNeuroModulatedHebbianLearning2023a}. Nevertheless, considering the average error rate across the three datasets, we believe that the proposed method can still offer certain benefits for simple transfer learning tasks such as digit recognition, without causing detrimental effects.

\subsection{Energy consumption}
\label{subsec:energy}

In addition to running error rate, we evaluated our method in terms of accumulate operations (ACs), multiply–accumulate operations (MACs), multiply operations (MULs) as well as the theoretical energy consumption of inference under the condition $\rho_{0}=1, r=0$ (see algorithm \ref{algo:tm}). It is worth noting that the energy consumption of SNNs running on actual neuromorphic chip is influenced by additional factors. Therefore, the estimation here is approximate.
\begin{itemize}
    \item \textit{MACs:} For layers with continuous inputs, such as the first convolutional layer, MAC operations are needed.
    \item \textit{ACs:} ACs are comprised of operations in the convolutional layers with spike input, LIF neurons and Threshold Modulation modules.
    \item \textit{MULs:} In the proposed Threshold Modulation modules, the computation of statistics and thresholds is composed of separate accumulation and multiplication operations. Therefore, we calculate ACs and MULs separately for this module.
    \item \textit{Energy estimation:} Based on MACs, ACs and MULs, we estimate the energy consumption during test-time of each sample in CIFAR-10-C with Gaussian Noise. The additional energy involved in backpropagation in TM-ENT will be discussed later in the ablation study. Energy consumption of each MAC, ACC or MUL operation is based on the $45nm$ hardware \cite{horowitz11ComputingsEnergy2014}, where each AC costs $0.9pJ$, each MUL costs $3.7pJ$ and each MAC cost $4.6pJ$.
\end{itemize}
\begin{equation}
    \mathrm{SOPs}^{l}=fr^{l-1}\times T\times \mathrm{MACs^\mathit{l}}\label{eq:sop}
\end{equation}
\begin{equation}
\begin{split}
    E_{Total} &=E_{MAC}\times\mathrm{MACs_{Conv1}}\\
    &+E_{AC}\times(\sum\nolimits_{i}\mathrm{ACs_{TM}^{\mathit{i}}}
    +\sum\nolimits_{j}\mathrm{SOPs_{Conv}^\mathit{j}}\\
    &+\mathrm{ACs_{FC}}+\sum\nolimits_{k}\mathrm{ACs_{LIF}^\mathit{k}}) \\
    &+E_{MUL}\times \sum\nolimits_{i}\mathrm{MULs_{TM}^{\mathit{i}}}
    \label{eq:energy}
\end{split}
\end{equation}

Equation~\eqref{eq:sop} defines the number of Synaptic Operations (SOPs), where $l$ is a network layer with spike input. $fr$ is the average firing rate of the spike input and $T(=4)$ is the number of simulation time steps, $E_{SOP}=E_{AC}$. Equation~\eqref{eq:energy} calculates the total theoretical energy consumption of the SNN models used in this work, where $\mathrm{TM}$ denotes the Threshold Modulation module, $\mathrm{FC}$ denotes the final classification layer, $\mathrm{Conv1}$ denotes the input convolutional layer, $\mathrm{Conv}$ denotes the spiking convolutional layers and $\mathrm{LIF}$ denotes the LIF neurons. Following \cite{duanBrainInspiredOnlineAdaptation2025}, each LIF neuron requires 1 AC per time step to update its membrane potential. The results are summarized in Tab.~\ref{tab:energy}. The results show that, when pre-training with MPBN, using TM-NORM reduces energy consumption by lowering the overhead of computing statistics during inference compared to directly calibrating the BN, with even a slight decrease compared with pure inference. Compared to the model without MPBN (without adaptation ability), TM-NORM only increases power consumption by 3\%, which is much lower than the additional 12\% required for direct calibration of MPBN. Therefore, the proposed TM-NORM method greatly improves the robustness of the model without causing much increase in energy consumption.

\begin{table}[t]
    \setlength{\abovecaptionskip}{0pt}
    \caption{Theoretical Energy Consumption per sample on CIFAR-10-C with Gaussian Noise using Spiking VGG-16m ($\mathrm{T}=4$).}
    \centering
    \begin{tabular}{@{\hskip 3pt}l@{\hskip 4pt}|@{\hskip 4pt}c@{\hskip 4pt}|@{\hskip 4pt}c@{\hskip 4pt}|@{\hskip 4pt}c@{\hskip 4pt}|@{\hskip 4pt}l@{\hskip 3pt}l}
    \toprule
    Method & MACs & ACs & MULs & Energy ($\mu\text{J}$)\\
    \midrule
    w/o MPBN & 1.84M & 177.33M & 0.00M & \underline{168.04} \\
    MPBN (pure inference) & 1.84M & 175.99M & 2.21M & 175.01(+4\%)\\
    MPBN (direct calibration) & 1.84M & 186.21M & 3.35M & 188.43(+12\%) \\ 
    TM (pure inference) & 1.84M & 203.27M & 0.03M & 191.51(+14\%) \\
    TM-NORM & 1.84M & 182.11M & 0.26M & \textbf{173.29(+3\%)}\\
    \bottomrule
    \end{tabular}
    \label{tab:energy}
    \vspace{-12pt}
\end{table}

\subsection{Ablation study}
\subsubsection{Influence of momentum-based update of statistics and normalization of non-firing potentials} 
When estimating the mean and variation of membrane potentials, momentum-based updates can be employed (by setting $\rho_{0}<1$). In order for the full equivalence of the re-parameterized model, normalization of non-firing neurons can be applied (by setting $r=1$). As described in \ref{subsec:results}, we use none of them in the adaptation to common corruptions task but both in digit recognition transfer task. However, either approach will introduce additional energy consumption. Here, we compare the situation with or without momentum-based updates and residual potential normalization.
\subsubsection{Influence of entropy minimization}
\begin{figure}[tbp]
    \centering
    \begin{minipage}{0.95\linewidth}
        \raggedleft
        \includegraphics[width=\linewidth]{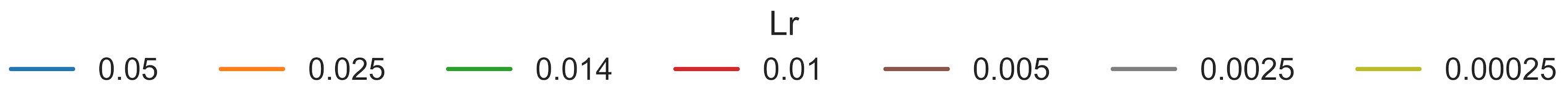}
        \label{subfig:ablation_legend}
    \end{minipage}%
    \vspace{-\baselineskip}
    \begin{minipage}{0.5\linewidth}
        \centering
        \includegraphics[width=\linewidth]{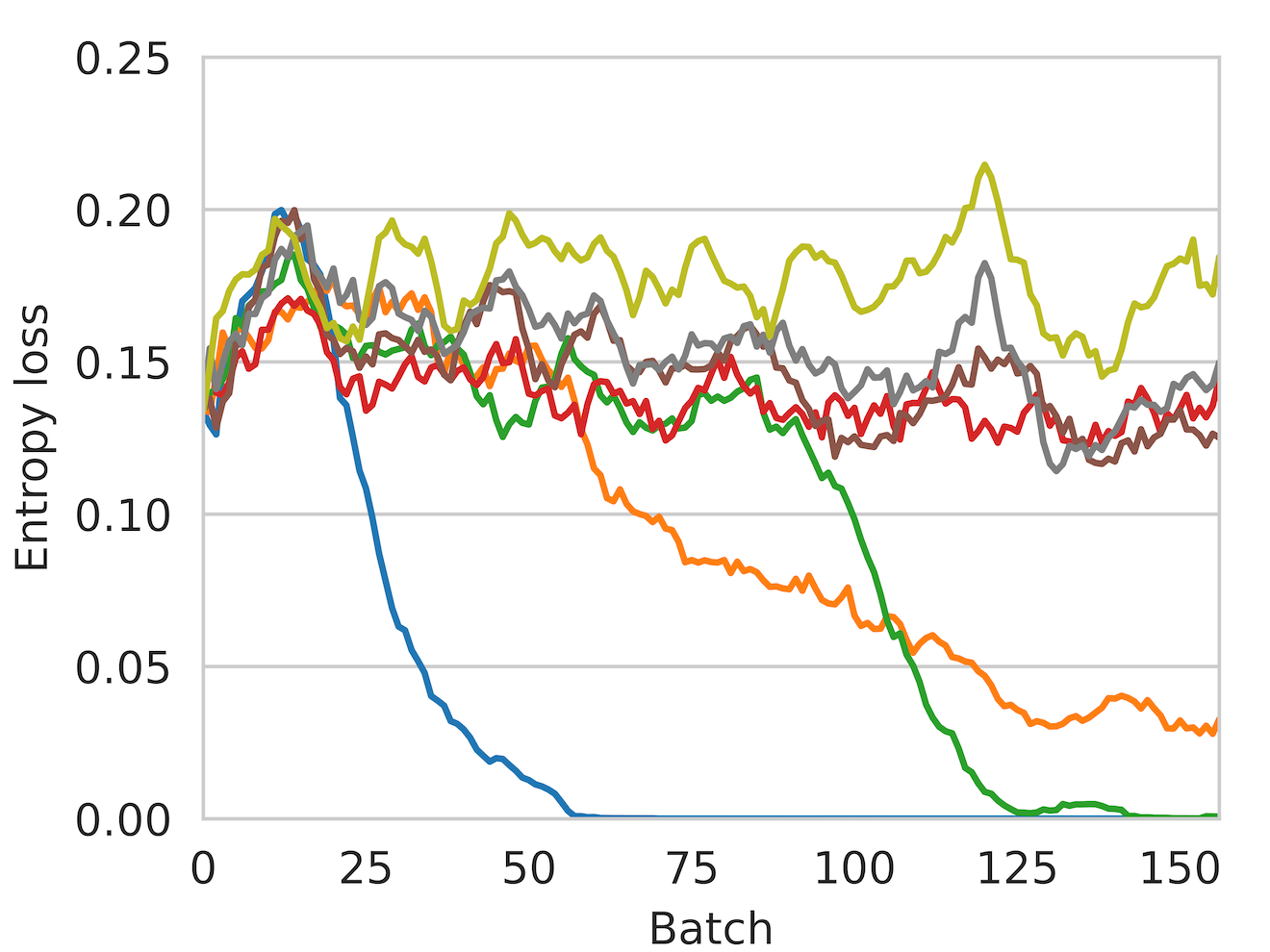}
        \label{subfig:ablation_loss}
    \end{minipage}%
    \begin{minipage}{0.5\linewidth}
        \centering
        \includegraphics[width=\linewidth]{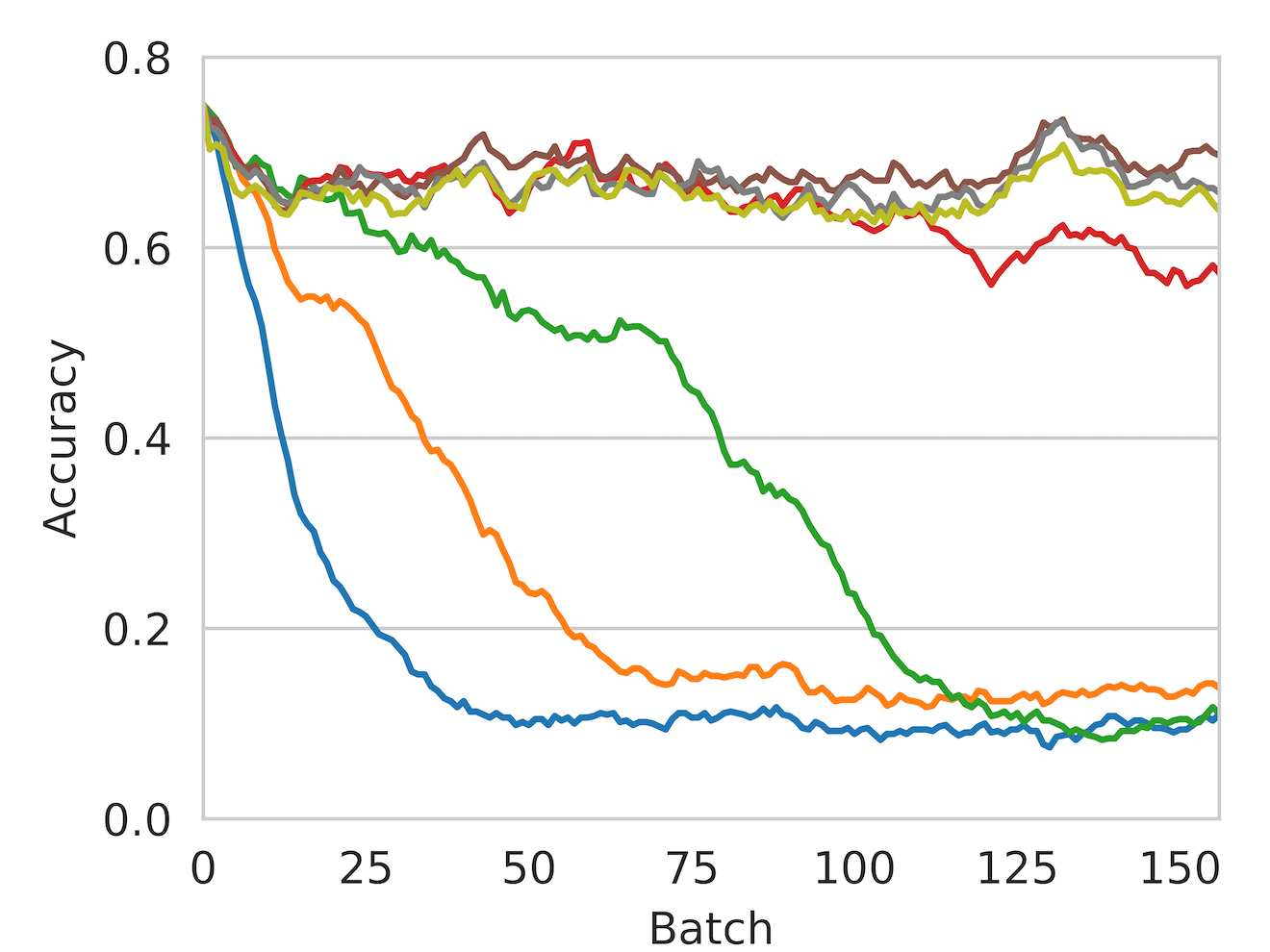}
        \label{subfig:ablation_accuracy}
    \end{minipage}%
    \vspace{-\baselineskip}
    \begin{minipage}{0.5\linewidth}
        \centering
        \includegraphics[width=\linewidth]{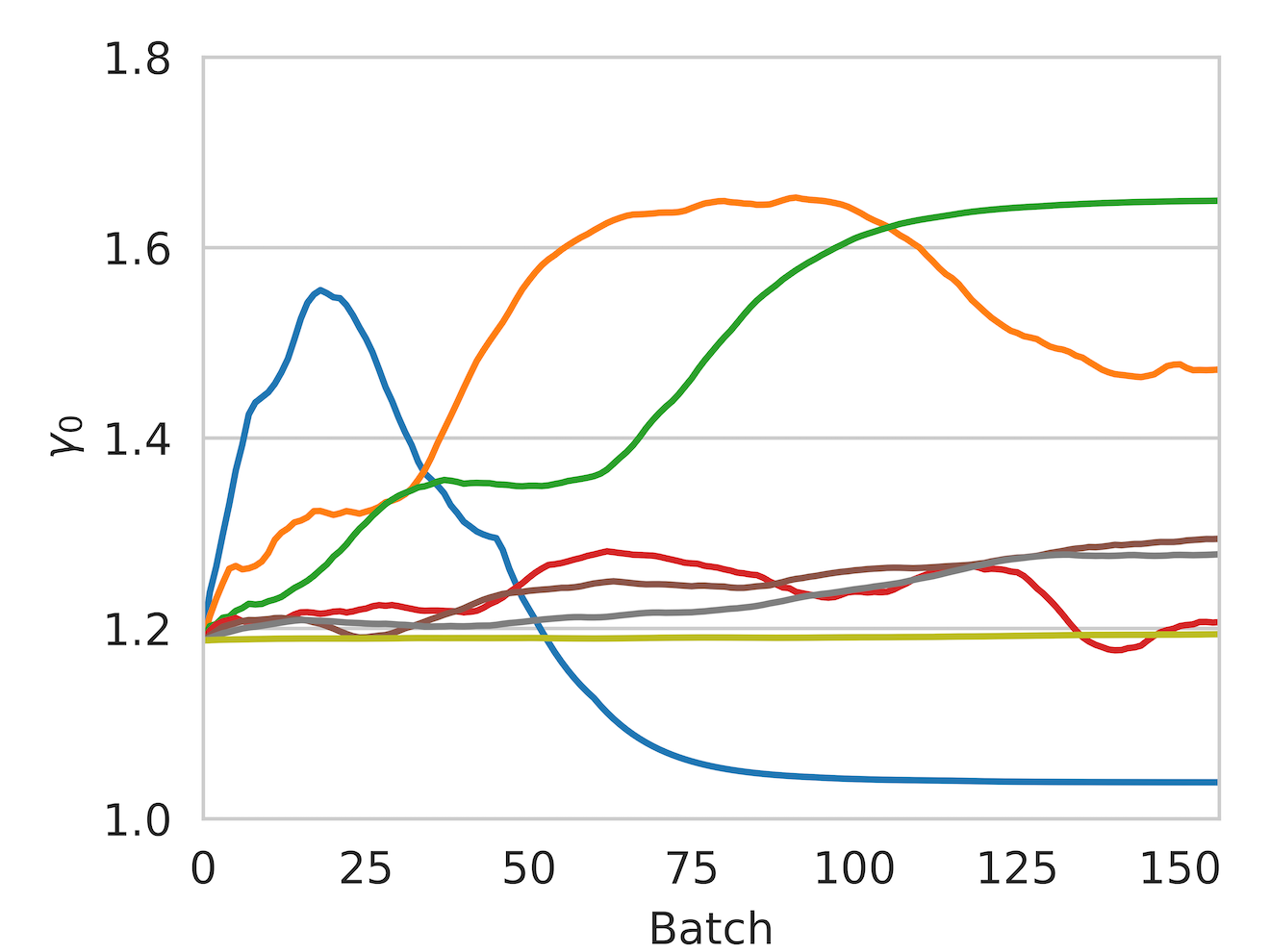}
        \label{subfig:ablation_gamma}
    \end{minipage}%
    \begin{minipage}{0.5\linewidth}
        \centering
        \includegraphics[width=\linewidth]{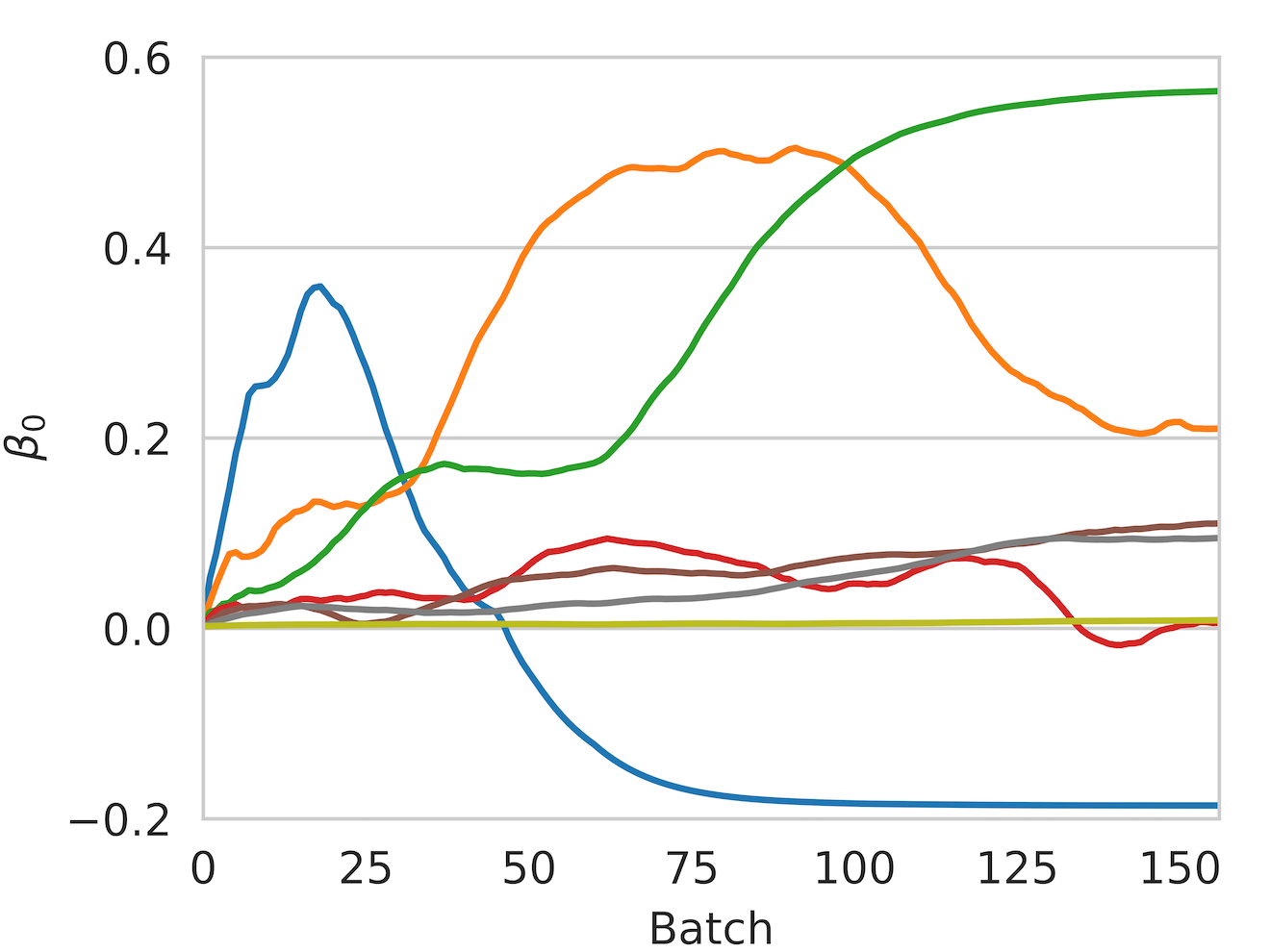}
        \label{subfig:ablation_beta}
    \end{minipage}%
    \vspace{-4pt}
    \caption{Test entropy, running accuracy and the value of affine parameters as the online batch input.}
    \label{fig:ablation}
    \vspace{-10pt}
\end{figure}

Entropy minimization is widely used in online test-time adaptation for ANNs, while some work pointed out the vulnerabilities of the vanilla version as it may result in collapsed trivial solutions \cite{boudiafParameterfreeOnlineTesttime2022,niuStableTesttimeAdaptation2022,leeEntropyNotEnough2023}. TM-NORM sometimes outperforms TM-ENT although it has no trainable parameters. Fig.~\ref{fig:ablation} also shows that choosing a proper learning rate without leading to model collapsing is quite tricky, and the affine parameters are likely to change dramatically (data points were smoothed). Most importantly, employing this method in the online adaptation of SNNs necessitates the use of back-propagation or other optimization algorithms, which will significantly increase energy consumption and introduce additional challenges for on-chip implementation. Even if we use online training algorithms like SLTT \cite{mengMemoryTimeEfficientBackpropagation2023a} to update the affine parameters, when accounting for the computational overhead introduced by back-propagation, the computational cost of TM-ENT will be multiple times greater than that of the purely feed-forward TM-NORM (like models V1 and V2 in Tab. \ref{tab:ablation}), let alone additional storage requirement. Here, the energy estimation of back-propagation is based on SLTT-4, and the other model weights are frozen.
\subsubsection{Results} The ablation study considered various combinations of $\{\rho_{0}, r, e\}$ (see algorithm.~\ref{algo:tm}) and compared their accuracy and energy consumption. The results in Tab.~\ref{tab:ablation} provide insights into selecting the most efficient combination. Based on our experiments and analysis, we find that entropy minimization has limited significance while greatly increase the energy consumption. Normalizing the membrane potential of non-firing neurons helps improve accuracy; however, even without this operation, while it is not strictly equivalent to the model without re-parameterization, it does not lead to a significant accuracy drop. Overall, in practical applications, the components to be enabled can be selected strictly based on energy consumption and computational capacity. For example, V5 can be selected for most simplicity, and V2 can be selected for higher performance.

\begin{table}[t]
    \setlength{\abovecaptionskip}{0pt}
    \caption{Running Accuracy and Energy Estimation on CIFAR-10-C \\ with Gussian Noise using Spiking VGG-16m.}
    \centering
    \begin{tabular}{l|c|c|c|c|c}
        \toprule
        Model & $\rho_{0}<1$ & $r=1$ & $e=1$ & Acc. (\%) & Energy ($\mu\text{J}$)\\
        \midrule
        V1 & \checkmark & \checkmark & \checkmark & \textbf{69.05} & 1316.51\\
        V2 & \checkmark & \checkmark & $\times$ & 68.64 & 180.88 \\
        V3 & $\times$ & \checkmark & $\times$ & 68.57 & 183.73 \\
        V4 & \checkmark & $\times$ & $\times$ & 66.48 & 174.79\\
        V5 & $\times$ & $\times$ & $\times$ & 67.60 & \textbf{173.29}\\
        \bottomrule
    \end{tabular}
    \label{tab:ablation}
    \vspace{-12pt}
\end{table}

\section{Discussion}

\subsection{Spiking neuron models with adaptive threshold}
The proposed Threshold Modulation (TM) module dynamically adjusts the threshold $V_{th}$ used for neuronal firing in the current time step based on normalization calibration, aiming to correct internal covariate shifts. On the other hand, neuron models with adaptive threshold \cite{bellecSolutionLearningDilemma2020,bellecLongShorttermMemory2018} dynamically adjust the threshold based on spike activity, while learnable thresholds \cite{wangLTMDLearningImprovement2022,rathiDIETSNNLowLatencySpiking2023} update $V_{th}$ with gradient descent to approximate neuronal dynamics. Here, we compare the performance of the proposed TM module with channel-wise learnable threshold and the Adaptive LIF \cite{bellecSolutionLearningDilemma2020} on CIFAR-10-C. All networks were trained from scratch using the same hyper-parameters as \ref{subsec:implementation}; the learnable thresholds are updated by entropy minimization during adaptation. The Results on CIFAR-10-C are shown in Tab.~\ref{tab:discussion}: neither using the adaptive LIF nor learning threshold by entropy minimization helps improve the accuracy on the corrupted dataset; direct training with the Adaptive LIF suffers from over-fitting, and the learnable threshold does not help in this case either.

\begin{table}[t]
    \setlength{\abovecaptionskip}{0pt}
    \caption{Comparison of different adaptive threshold method on CIFAR-10-C with Gaussian Noise using Spiking ResNet-20.}
    \centering
    \begin{tabular}{l|c|c|c}
        \toprule
        Method / Acc. (\%) & Clean & Source & Adaptation \\
        \midrule
        Adaptive LIF & 63.92 & 31.62 & $-$ \\
        Learnable threshold & 94.55 & 51.86 & 49.25 (-2.61)\\
        TM-NORM (ours) & 93.04 & 53.28 & 74.82 (\textbf{+21.54}) \\
        \bottomrule
    \end{tabular}
    \label{tab:discussion}
    \vspace{-12pt}
\end{table}

\subsection{Limitations and future works}
The proposed TTA framework for SNNs aims to integrate three stages: pre-training, deployment, and online test-time adaptation. However, this work also has certain limitations. It must be acknowledged that the current implementation relies on the MPBN module and has not been extensively tested for compatibility with other architectures and SNN training techniques. For example, convergence issues may arise. Additionally, the evaluation of the proposed method has been conducted on a limited set of datasets, which may not fully capture the diversity of real-world scenarios. Further experiments on more diverse and complex datasets are necessary to validate the generalizability of the proposed approach. When deploying on actual neuromorphic chips, it is necessary to consider the compatibility of the specific chip with statistical computation and entropy minimization. Unsupported operators may introduce unforeseen energy consumption.

Moreover, while MPBN is used in the current implementation, alternative approaches for integrating TM into SNNs remain an open direction for further exploration. For examples, referring to \cite{esserConvolutionalNetworksFast2016} for mapping CNNs onto neuromorphic chips, a leakage term  $L=\left \lceil \beta(\sigma+\epsilon )-\mu \right \rceil$ is introduced into the charging function of the Integrate-and-Fire (IF) model, where $L$ is determined by the pre-trained BN layer. If the values of $\mu, \sigma$ can be adjusted based on the input statistics, the membrane potential distribution could be regulated by modifying the leakage term instead.

Despite these limitations, our work serves as one of the pioneering studies in test-time adaptation for SNNs, demonstrating the feasibility of Threshold Modulation in this context. The successful application of TM in our framework highlights its potential for enhancing the on-chip adaptability of SNNs to distribution shifts, thereby laying the foundation for future research. Future research could extend this framework by incorporating more advanced SNN pretraining techniques to further
enhance performance. It could also be tested and optimized
for broader scenarios such as continuous online adaptation, as well as tasks other than image classification, drawing insights from state-of-the-art techniques for ANNs.

\section{Conclusion}

This work presents a low-power, neuromorphic chip-friendly online test-time adaptation framework for SNNs, being one of the first works to address this issue. Experimental results on benchmark datasets demonstrate that the proposed method can effectively help models deployed on neuromorphic hardware handle distribution shifts. Despite some limitations, this method can serve as a new guide for future SNN online test-time adaptation research and neuromorphic chip design.

\section{Acknowledgment}
This research was partially supported by the National Natural Science Foundation of China (No. 62202217), Guangdong Basic and Applied Basic Research Foundation (No. 2023A1515012889), Guangdong Key Program (No. 2021QN02X794), and STI 2030-Major Projects 2022ZD0208700.

\bibliographystyle{IEEEtran}
\bibliography{references}

\end{document}